# Boolean Equi-propagation for Concise and Efficient SAT Encodings of Combinatorial Problems

**Amit Metodi** AMIT.METODI@GMAIL.COM
**Michael Codish** MCODISH@CS.BGU.AC.IL
*Department of Computer Science*
*Ben-Gurion University of the Negev, Israel*

**Peter J. Stuckey** PJS@CSSE.UNIMELB.EDU.AU
*Department of Computer Science and Software Engineering*
*and NICTA Victoria Laboratory*
*The University of Melbourne, Australia*

## Abstract

We present an approach to propagation-based SAT encoding of combinatorial problems, Boolean equi-propagation, where constraints are modeled as Boolean functions which propagate information about equalities between Boolean literals. This information is then applied to simplify the CNF encoding of the constraints. A key factor is that considering only a small fragment of a constraint model at one time enables us to apply stronger, and even complete, reasoning to detect equivalent literals in that fragment. Once detected, equivalences apply to simplify the entire constraint model and facilitate further reasoning on other fragments. Equi-propagation in combination with partial evaluation and constraint simplification provide the foundation for a powerful approach to SAT-based finite domain constraint solving. We introduce a tool called BEE (Ben-Gurion Equi-propagation Encoder) based on these ideas and demonstrate for a variety of benchmarks that our approach leads to a considerable reduction in the size of CNF encodings and subsequent speed-ups in SAT solving times.

## 1. Introduction

In recent years, Boolean SAT solving techniques have improved dramatically. Today's SAT solvers are considerably faster and able to manage larger instances than yesterday's. Moreover, encoding and modeling techniques are better understood and increasingly innovative. SAT is currently applied to solve a wide variety of hard and practical combinatorial problems, often outperforming dedicated algorithms. The general idea is to encode a (typically, NP) hard problem instance, $\mu$, to a Boolean formula, $\varphi_\mu$, such that the satisfying assignments of $\varphi_\mu$ correspond to the solutions of $\mu$. Given such an encoding, a SAT solver can be applied to solve $\mu$.

Tailgating the success of SAT technology are a variety of tools which can be applied to specify and then compile problem instances to corresponding SAT instances. The general objective of such tools is to facilitate the process of providing high-level descriptions of how the (constraint) problem at hand is to be solved. Typically, a constraint-based modeling language is introduced and used to model instances. Drawing on the analogy to programming languages, given such a description, a compiler can then provide a low-level executable for the underlying machine. Namely, in our context, a formula for the underlying SAT or SMT solver.

For example, Cadoli and Schaerf (2005) introduce NP-SPEC, a logic-based specification language which allows specifying combinatorial problems in a declarative way. At the core of this system is a component which translates specifications to CNF formula. Similarly Sugar (Tamura, Taga, Kitagawa, & Banbara, 2009) is a SAT-based constraint solver. To solve a finite domain constraint satisfaction problem it is first modeled in a constraint language (also called Sugar) and then encoded to a CNF formula and solved using the MiniSAT solver (Eén & Sörensson, 2003). MiniZinc (Nethercote, Stuckey, Becket, Brand, Duck, & Tack, 2007) is a constraint modeling language that is compiled by a variety of solvers to the low-level target language FlatZinc for which there exist many solvers. In particular, FlatZinc instances are solved by `fzntini` (Huang, 2008) by encoding them to CNF and in `fzn2smt` by encoding to SMT-LIB (Barrett, Stump, & Tinelli, 2010).

Simplifying CNF formulae prior to the application of SAT solving is of the utmost importance and there are a wide range of techniques that can be applied to achieve this goal. See for example the work of Li (2003), Eén and Biere (2005), Heule, Järvisalo, and Biere (2011), and Manthey (2012), and the references therein their work. All of these techniques exhibit a clear trade-off between the amount of simplification obtained and the time it requires. Moreover, the stronger techniques become prohibitive when the SAT model involves hundreds of thousands of variables and millions of clauses. So in CNF simplification tools, time limits on simplification techniques are imposed and/or approximations are used.

This paper takes a new approach to CNF simplification. Typically, a CNF is not a random collection of clauses, but rather has a structure derived from an application or specific problem domain. When SAT solving is applied to encode and solve finite domain constraint problems, the original constraint model is a manifest of this structure. Usually, the constraints are discarded once encoded to CNF. We advocate that maintaining the constraints provides important structural information that can be applied to drive the process of CNF simplification. To be specific, the constraints in a model induce a partitioning of their CNF encoding to a conjunction of sub-formulae which we call "portions".

The novelty in our approach to CNF simplification is that instead of considering the CNF as a whole, we assume that it is partitioned into a conjunction of smaller portions. Then simplification is repeatedly applied to individual portions. This facilitates a propagation-based process because the simplification of one portion propagates information to all of the portions and this information may trigger further simplification in other portions.

Because portions are typically much smaller than the entire CNF we can effectively apply stronger simplification algorithms. We introduce the notion of equi-propagation. Similar to how unit propagation is about inferring unit clauses which can then be applied to simplify CNF formulae, equi-propagation is about inferring equational consequences between literals (and Boolean constants).

There is a wide body of research on CNF simplification that can be applied to implement equi-propagation which is sometimes called equivalent literal substitution, for example by Gelder (2005). Techniques typically involve binary clause based simplifications using, among others, hyper binary resolution and binary implication graphs. See for example, the work of Heule et al. (2011) and the references therein. The guiding principle in all of these works is that techniques must be simple and efficient because of the prohibitive size of the CNF to which they must apply.

Our approach is different and we focus on far richer forms of inference not even related to the CNF structure of a formula. At one extreme we apply complete equi-propagation which detects all equivalences implied by a formula. Clearly complete equi-propagation is NP-hard. However, complete equi-propagators are feasible as we apply them only to small portions of the formula. When complete equi-propagation is too slow we consider ad-hoc techniques. All of these forms of equi-propagation have in common that they are not driven by the CNF structure (e.g. binary clauses) but rather by the underlying constraint structure from which a CNF was, or is being, generated.

The rest of this paper is structured as follows. Section 2 introduces a modeling language for finite domain constraints which consists of just 5 constraint constructs and is sufficient to illustrate the contribution of the paper. We argue that the constraints in a model induce a natural partition of their CNF encoding to smaller portions and that this partition can be used to drive the simplification of the CNF encoding. Section 3 presents equi-propagation which is the first ingredient for our contribution. Equi-propagation is about learning information that will apply to simplify CNF encodings. Section 4 describes a practical basis for implementing equi-propagation. Section 5 introduces the second ingredient: partial evaluation. Given the information derived using equi-propagation, partial evaluation applies to simplify the constraints and in particular to remove Boolean variables from their CNF encodings. Section 6 describes a tool, called BEE (Metodi & Codish, 2012) (Ben-Gurion Equi-propagation Encoder) that is based on equi-propagation and partial evaluation. We introduce here our full constraint language which is similar to Sugar and to the subset of FlatZinc relevant for finite domain constraint problems. We also spell out the special treatment of the all-different constraint in BEE. Section 7 demonstrates the application of BEE. Section 8 presents an experimental evaluation. and Finally Section 9 presents our conclusion.

This paper extends earlier work presented by Metodi, Codish, Lagoon, and Stuckey (2011), which first introduced equi-propagation, and also the BEE tool paper (Metodi & Codish, 2012). The BEE tool is available for download (Metodi, 2012).

## 2. Constraint Based Boolean Modeling

This section provides the basis for our contribution: a constraint-based modeling language, together with a Boolean interpretation for each constraint in the language. This enables us to view a constraint model as a conjunction of Boolean formulae and provides a structure which drives the subsequent encoding to CNF.

We first introduce a simple and small fragment of a typical finite domain constraint-based modeling language. This serves to illustrate our approach. Later, in Section 6, we show the full language. We then discuss several options for Boolean representation of integers. In this paper we adopt a particular unary representation, called the order encoding. Our contribution is independent of this choice, although equi-propagation works well with it. Finally we finish the section so that each of the constraints in the language fragment can be viewed as a Boolean formula, and a constraint model as their conjunction.

| | | |
|---|---|---|
| (1) | $\mathtt{new_int}(\mathtt{I}, \mathtt{c_1}, \mathtt{c_2})$ | $0 \leq \mathtt{c_1} \leq \mathtt{I} \leq \mathtt{c_2}$ |
| (2) | $\mathtt{int_neq}(\mathtt{I_1}, \mathtt{I_2})$ | $\mathtt{I_1} \neq \mathtt{I_2}$ |
| (3) | $\mathtt{allDiff}([\mathtt{I_1}, \ldots, \mathtt{I_n}])$ | $\bigwedge_{\mathtt{i<j}} \mathtt{I_i} \neq \mathtt{I_j}$ |
| (4) | $\mathtt{int_plus}(\mathtt{I_1}, \mathtt{I_2}, \mathtt{I})$ | $\mathtt{I_1} + \mathtt{I_2} = \mathtt{I}$ |
| (5) | $\mathtt{int_array_plus}([\mathtt{I_1}, \ldots, \mathtt{I_n}], \mathtt{I})$ | $\mathtt{I_1} + \cdots + \mathtt{I_n} = \mathtt{I}$ |

Figure 1: A core constraint language

### 2.1 Constraint Language Fragment

We focus on a small fragment of a typical constraint modeling language detailed in Figure 1. This serves to present the main ideas of the paper. Constraint (1) is about declaring finite domain integer variables in the range $[c_1...c_2]$. For simplicity in the presentation we will further assume that $c_1 \geq 0$. Constraints (2–3) are about difference of integer variables, and constraints (4–5) are about sums of integer variables. As syntactic sugar we also allow writing integer constants in constraints. For example, $\mathtt{int_neq}(\mathtt{I}, 5)$ which is short for $\mathtt{new_int}(\mathtt{I'}, 5, 5)$, $\mathtt{int_neq}(\mathtt{I}, \mathtt{I'})$.

### 2.2 Modeling Kakuro: an Example

A Kakuro puzzle is an $n \times m$ board of black and white cells. The black cells contain hints and the white cells are to be filled by numbers between 1 and 9 (the bound 9 is often generalized by a larger value $r$). The hints specify constraints on the sums of the values in blocks of white cells to the right and/or below the hint. The numbers assigned to the white cells in such a block are required to be "all different". Figure 2 illustrates a $4 \times 4$ Kakuro puzzle (left) and its solution (right).

To model a Kakuro puzzle we view it as a set of blocks (of white cells) where each block $B$ is a set of integer variables and is associated with a corresponding integer value, $hint(B)$. Each block $B$ is associated with two constraints: the integers in $B$ must sum to $hint(B)$ and must be all-different. Figure 3 illustrates the constraints corresponding to the Kakuro instance in Figure 2.

### 2.3 Representing Integers

A fundamental design choice when encoding finite domain constraints concerns the representation of integer variables. Gavanelli (2007) surveys several of the possible choices (the

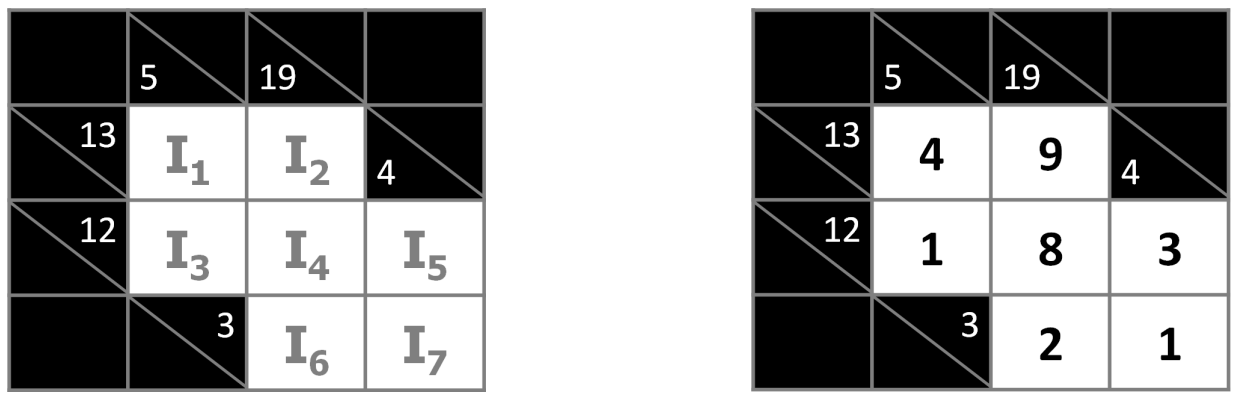

Figure 2: A $4 \times 4$ Kakuro puzzle (right) and its solution (left).

| | | |
|---|---|---|
| $\texttt{new_int}(\texttt{I}_1, 1, 9)$ | $\texttt{int_array_plus}([\texttt{I}_1, \texttt{I}_2], 13)$ | $\texttt{allDiff}([\texttt{I}_1, \texttt{I}_2])$ |
| $\texttt{new_int}(\texttt{I}_2, 1, 9)$ | $\texttt{int_array_plus}([\texttt{I}_1, \texttt{I}_3], 5)$ | $\texttt{allDiff}([\texttt{I}_1, \texttt{I}_3])$ |
| $\texttt{new_int}(\texttt{I}_3, 1, 9)$ | $\texttt{int_array_plus}([\texttt{I}_3, \texttt{I}_4, \texttt{I}_5], 12)$ | $\texttt{allDiff}([\texttt{I}_3, \texttt{I}_4, \texttt{I}_5])$ |
| $\texttt{new_int}(\texttt{I}_4, 1, 9)$ | $\texttt{int_array_plus}([\texttt{I}_2, \texttt{I}_4, \texttt{I}_6], 19)$ | $\texttt{allDiff}([\texttt{I}_2, \texttt{I}_4, \texttt{I}_6])$ |
| $\texttt{new_int}(\texttt{I}_5, 1, 9)$ | $\texttt{int_array_plus}([\texttt{I}_6, \texttt{I}_7], 3)$ | $\texttt{allDiff}([\texttt{I}_6, \texttt{I}_7])$ |
| $\texttt{new_int}(\texttt{I}_6, 1, 9)$ | $\texttt{int_array_plus}([\texttt{I}_5, \texttt{I}_7], 4)$ | $\texttt{allDiff}([\texttt{I}_5, \texttt{I}_7])$ |
| $\texttt{new_int}(\texttt{I}_7, 1, 9)$ | | |

Figure 3: Constraints for the Kakuro instance of Figure 2.

*direct-*, *support-* and *log- encodings*) and introduces the *log-support encoding.* Given a choice of representation constraints are bit-blasted and interpreted as Boolean formulae. We focus for now on the use a unary representation, the so-called, *order-encoding* (see, e.g. Crawford & Baker, 1994; Bailleux & Boufkhad, 2003) which has many nice properties when applied to small finite domains.

In the *order-encoding*, an integer variable $X$ in the domain $[0,\ldots,n]$ is represented by a bit vector $X = [x_1,\ldots,x_n]$. Each bit $x_i$ is interpreted as $X \geq i$ so in particular the bit sequence $X$ constitutes a monotonic non-increasing Boolean sequence. For example, the value 3 in the interval $[0,5]$ is represented in 5 bits as $[1,1,1,0,0]$.

An important property of a Boolean representation for finite domain integers is the ability to represent changes in the set of values a variable can take. It is well-known that the order-encoding facilitates the propagation of bounds. Consider an integer variable $X = [x_1,\ldots,x_n]$ with values in the interval $[0,n]$. To restrict $X$ to take values in the range $[a,b]$ (for $1 \leq a \leq b \leq n$), it is sufficient to assign $x_a = 1$ and $x_{b+1} = 0$ (if $b < n$). The variables $x_{a'}$ and $x_{b'}$ for $0 \geq a' > a$ and $b < b' \leq n$ are then determined true and false, respectively, by *unit propagation.* For example, given $X = [x_1,\ldots,x_9]$, assigning $x_3 = 1$ and $x_6 = 0$ propagates to give $X = [1,1,1,x_4,x_5,0,0,0,0]$, signifying that $dom(X) = \{3,4,5\}$.

We observe an additional property of the order-encoding for $X = [x_1,\ldots,x_n]$: its ability to specify that a variable cannot take a specific value $0 \leq v \leq n$ in its domain by equating two variables: $x_v = x_{v+1}$. This indicates that the order-encoding is well-suited not only to propagate lower and upper bounds, but also to represent integer variables with an arbitrary, finite set, domain. For example, given $X = [x_1,\ldots,x_9]$, equating $x_2 = x_3$ imposes that $X \neq 2$. Likewise $x_5 = x_6$ and $x_7 = x_8$ impose that $X \neq 5$ and $X \neq 7$. Applying these equalities to $X$ gives, $X = [x_1,\underline{x_2,x_2},x_4,\underline{x_5,x_5},\underline{x_7,x_7},x_9]$ (note the repeated literals), signifying that $dom(X) = \{0,1,3,4,6,8,9\}$.

The order-encoding has many additional nice features that can be exploited to simplify constraints and their encodings to CNF. To illustrate one, consider a constraint of the form $\texttt{A} + \texttt{B} = \texttt{5}$ where $\texttt{A}$ and $\texttt{B}$ are integer values in the range between 0 and 5 represented in the order-encoding. At the bit level (in the order encoding) we have: $\texttt{A} = [\texttt{a}_1,\ldots,\texttt{a}_5]$ and $\texttt{B} = [\texttt{b}_1,\ldots,\texttt{b}_5]$. The constraint is satisfied precisely when $\texttt{B} = [\neg\texttt{a}_5,\ldots,\neg\texttt{a}_1]$. Instead of encoding the constraint to CNF, we substitute the bits $\texttt{b}_1,\ldots,\texttt{b}_5$ by the literals $\neg\texttt{a}_5,\ldots,\neg\texttt{a}_1$, and remove the constraint. In section 3 we formalize this process of discovering equalities between literals implied by a constraint and using them to simplify CNF encodings.

### 2.4 Bit Blasting

Given a constraint model and the decision on how to represent finite domain integer variables at the bit level (we chose the order encoding), "bit-blasting" is the process of instantiating integer variables by corresponding bit vectors and interpreting constraints as Boolean formulae.

Each integer variable, $\mathtt{I}$, declared by a constraint of the form $\mathtt{new_int}(\mathtt{I}, \mathtt{c_1}, \mathtt{c_2})$ where $0 \leq c_1 \leq c_2$ is represented as a bit-vector $\mathtt{I} = [\mathtt{1}, \ldots, \mathtt{1}, \mathtt{X_{c_1+1}}, \ldots, \mathtt{X_{c_2}}]$. So, we may view a constraint model as consisting only of Boolean variables and each constraint $c$ corresponds to a Boolean formula denoted as $[\![c]\!]$, the "bit-blasted" version of $c$. The specific definition of $[\![\cdot]\!]$ is not important. Just for illustration, note that one could define

$$[\![\mathtt{new_int}(\mathtt{I}, \mathtt{c_1}, \mathtt{c_2})]\!] = \bigwedge_{c_1 \leq i < c_2} (x_{i+1} \rightarrow x_i)$$

where $I = [1, \ldots, 1, X_{c_1+1}, \ldots, X_{c_2}]$ as well as

$$[\![\mathtt{int_neq}(\mathtt{I_1}, \mathtt{I_2})]\!] = \bigvee_{i=1}^{n} (x_i \ \mathtt{xor}\ y_i)$$

where to simplify presentation we assume that $I_1 = [x_1, \ldots, x_n]$ and $I_2 = [y_1, \ldots, y_n]$ are represented in the same number of bits. The mapping $[\![\cdot]\!]$ extends in the natural way to apply to conjunctions of constraints. So, given a constraint model such as the one in Figure 3, integer variables are instantiated to unary (order encoding) bit vectors and each constraint is viewed as a Boolean formula. The constraint model takes a Boolean representation as the conjunction of these formulae.

## 3. Boolean Equi-propagation

In this section we present an approach to propagation-based SAT encoding, Boolean equi-propagation, which propagates information about equalities between Boolean literals (and constants). We prove that Boolean equi-propagation is stronger than unit propagation as it determines at least as many fixed literals as unit propagation. We demonstrate, with an example, the power of equi-propagation and show that it leads to a considerable reduction in the size of the CNF encoding.

### 3.1 Boolean Equi-propagation

Let $\mathcal{B}$ be a set of Boolean variables. A *literal* is a Boolean variable $b \in \mathcal{B}$ or its negation $\neg b$. The negation of a literal $\ell$, denoted $\neg\ell$, is defined as $\neg b$ if $\ell = b$ and as $b$ if $\ell = \neg b$. The Boolean constants 1 and 0 represent *true* and *false*, respectively. The set of literals is denoted $\mathcal{L}$ and $\mathcal{L}_{0,1} = \mathcal{L} \cup \{0, 1\}$. The set of (free) Boolean variables that appear in a Boolean formula $\varphi$ is denoted $vars(\varphi)$. We extend the *vars* function to sets of formulae in the natural way.

An *assignment*, $A$, is a partial mapping from Boolean variables to constants, often viewed as the following set of literals: $\left\{\ b \mid A(b) = 1\ \right\} \cup \left\{\ \neg b \mid A(b) = 0\ \right\}$. For a formula $\varphi$ and $b \in \mathcal{B}$, we denote by $\varphi[b]$ (likewise $\varphi[\neg b]$) the formula obtained by substituting all occurrences of $b \in \mathcal{B}$ in $\varphi$ by *true* (*false*). This notation extends in the natural way for sets of literals. We say that $A$ satisfies $\varphi$ if $vars(\varphi) \subseteq vars(A)$ and $\varphi[A]$ evaluates to *true*. A *Boolean Satisfiability (SAT) problem* consists of a Boolean formula $\varphi$ and determines if there exists an assignment which satisfies $\varphi$.

A *Boolean equality* is a constraint $\ell = \ell'$ where $\ell, \ell' \in \mathcal{L}_{0,1}$. An *equi-formula* $E$ is a set of Boolean equalities understood as a conjunction. The set of Boolean equalities is denoted $\mathcal{L}_{0,1}^{eq}$ and the set of equi-formulae is denoted $\mathcal{E}$.

**Example 1.** *Suppose* $\mathcal{B} = \{x, y, z\}$. *Then* $\mathcal{L}_{0,1} = \{0, 1, \neg x, x, \neg y, y, \neg z, x\}$. *An example assignment is* $A = \{x, \neg z\}$*, while* $B = \{x, y, z, \neg y\}$ *is not an assignment (since it includes* $\{y, \neg y\}$*). Given the formula* $\varphi = x \leftrightarrow (y \vee \neg z)$ *then* $\varphi[\neg x]$ *is the formula* $0 \leftrightarrow (y \vee \neg z)$ *or equivalently* $\neg y \wedge z$. *The formula* $\varphi[A] = 1 \leftrightarrow (y \vee 1)$ *which is equivalent to true, but* $A$ *does not satisfy* $\varphi$ *since* $vars(\varphi) = \{x, y, z\} \not\subseteq \{x, z\} = vars(A)$. *An example equi-formula for* $\mathcal{B}$ *is* $\{x = 0, y = \neg z\}$ *or equivalently* $\neg x \wedge (y \leftrightarrow \neg z)$.

#### 3.1.1 Equi-propagation

is a process of inferring equational consequences from a Boolean formula and given equational information. An *equi-propagator* for a formula $\varphi$ is an extensive function $\mu_\varphi : \mathcal{E} \to \mathcal{E}$ defined such that for all $E \in \mathcal{E}$,

$$E \subseteq \mu_\varphi(E) \subseteq \left\{ e \in \mathcal{L}_{0,1}^{eq} \mid \varphi \wedge E \models e \right\}$$

That is, a conjunction of equalities, at least as strong as $E$, made true by $\varphi \wedge E$. We say that an equi-propagator $\mu_\varphi$ is complete if $\mu_\varphi(E) = \left\{ e \in \mathcal{L}_{0,1}^{eq} \mid \varphi \wedge E \models e \right\}$. We denote a complete equi-propagator for $\varphi$ as $\hat{\mu}_\varphi$. We assume that equi-propagators are monotonic: $E_1 \subseteq E_2 \Rightarrow \mu_\varphi(E_1) \subseteq \mu_\varphi(E_2)$. In particular, this follows, by definition, for complete equi-propagators. In Section 3.3 we discuss several methods to implement complete and incomplete equi-propagators.

**Example 2.** *Consider the constraint*

$$C = \texttt{new_int}(\texttt{X}, 0, 4) \ \wedge \ \texttt{new_int}(\texttt{Y}, 0, 4) \ \wedge \ \texttt{int_neq}(\texttt{X}, \texttt{Y})$$

*and its corresponding Boolean representation* $\varphi = [\![C]\!]$ *on the bit representation where*

$$X = [x_1, x_2, x_3, x_4] \text{ and } Y = [y_1, y_2, y_3, y_4]$$

*Assume the setting where*

$$E = \left\{ \ y_1 = 1, \ y_2 = 1, \ y_3 = 0, \ y_4 = 0 \ \right\}$$

*signifying that* $Y = 2$. *Then,* $\hat{\mu}_\varphi(E) = E \cup \{x_2 = x_3\}$ *indicating that* $X \neq 2$. *This occurs since* $\varphi \wedge E$ *is equivalent to* $(x_2 \to x_1) \wedge (x_3 \to x_2) \wedge (x_4 \to x_3) \wedge (\neg x_1 \vee \neg x_2 \vee x_3 \vee x_4)$ *and* $\varphi \wedge E \models x_2 = x_3$.

The following theorem states that complete equi-propagation is at least as powerful as unit propagation.

**Theorem 3.** *Let $\hat{\mu}_\varphi$ be a complete equi-propagator for a Boolean formula $\varphi$. Then, any literal that is made true by unit propagation for any clausal representation of $\varphi$ using the equations in $E$ is also determined true by $\hat{\mu}_\varphi(E)$.*

*Proof.* Let $\varphi$ be a Boolean formula, $E$ an equi-formula, and let $C_\varphi$ and $C_E$ be any clausal representations of $\varphi$ and of $E$ respectively. Clearly $\varphi \models C_\varphi$ and $E \models C_E$. Let $b$ be a positive literal determined by unit propagation of $C_\varphi \cup C_E$. Then by correctness of unit propagation, $C_\varphi \cup C_E \models b$. Hence, $\varphi \wedge E \models b$ and thus $\hat{\mu}_\varphi(E) \models b = 1$. The case for a negative literal $\neg b$ is the same, except that we infer $b = 0$. □

The following example illustrates that equi-propagation can be more powerful than unit propagation.

**Example 4.** *Consider $\varphi = (x_1 \leftrightarrow x_2) \wedge (x_1 \vee x_2) \wedge (\neg x_1 \vee \neg x_2 \vee \neg x_3)$. The clausal representation is $(x_1 \vee \neg x_2) \wedge (\neg x_1 \vee x_2) \wedge (x_1 \vee x_2) \wedge (\neg x_1 \vee \neg x_2 \vee \neg x_3)$ and no unit propagation is possible, since there are no unit clauses. Equi-propagation (with no additional equational information) gives: $\hat{\mu}_\varphi(\emptyset) = \{x_1 = 1, x_2 = 1, x_3 = 0\}$.*

### 3.1.2 Boolean Unifiers

It is sometimes convenient to view an equi-formula $E$ in a generic "solved-form" as a Boolean substitution, $\theta_E$, which is a (most general) unifier for the equations in $E$. Boolean substitutions generalize assignments in that variables can be bound also to literals. A Boolean *substitution* is an idempotent mapping $\theta : \mathcal{B} \to \mathcal{L}_{0,1}$ where $dom(\theta) = \left\{ b \in \mathcal{B} \,\middle|\, \theta(b) \neq b \right\}$ is finite. Note in particular that idempotence implies that $\theta(b) \neq \neg b$ for every $b \in \mathcal{B}$. Note also that $\theta$ is defined for all $\mathcal{B}$ and that its domain, $dom(\theta)$, includes those elements for which it is non-identity. A Boolean substitution, $\theta$, is viewed as the set $\theta = \left\{ b \mapsto \theta(b) \,\middle|\, b \in dom(\theta) \right\}$. We can apply $\theta$ to another substitution $\theta'$, to obtain substitution $(\theta \circ \theta') = \left\{ b \mapsto \theta(\theta'(b)) \,\middle|\, b \in dom(\theta) \cup dom(\theta') \right\}$. A *unifier* for equi-formula $E$ is a substitution $\theta$ such that $\models \theta(e)$, for each $e \in E$. A *most-general unifier* for $E$ is a substitution $\theta$ such that for any unifier $\theta'$ of $E$, there exists substitution $\gamma$ where $\theta' = \gamma \circ \theta$.

**Example 5.** Consider the equi-formula $E \equiv \{b_1 = \neg b_2, \neg b_3 = \neg b_4, b_5 = b_6, b_6 = b_4, b_7 = 1, b_8 = \neg b_7\}$ then a unifier $\theta$ for $E$ is $\{b_2 \mapsto \neg b_1, b_4 \mapsto b_3, b_5 \mapsto b_3, b_6 \mapsto b_3, b_7 \mapsto 1, b_8 \mapsto 0\}$. Note that $\theta(E)$ is the trivially true equi-formula $\{b_1 = \neg\neg b_1, \neg b_3 = \neg b_3, b_3 = b_3, b_3 = b_3, 1 = 1, 0 = \neg 1\}$.

Consider the enumeration $\mathcal{L}_{0,1} = \{0, 1, \neg b_1, b_1, \neg b_2, b_2, \ldots\}$ and let $\prec$ be the total (strict) order on $\mathcal{L}_{0,1}$ such that $0 \prec 1 \prec \neg b_1 \prec b_1 \prec \neg b_2 \prec b_2 \cdots$. We define a canonical most-general unifier $\texttt{unify}_E$ for any satisfiable equi-formula $E$ where:

$$\texttt{unify}_E(b) = \min \left\{ \ell \in \mathcal{L}_{0,1} \,\middle|\, E \models b = \ell \right\}$$

That is, the substitution $\texttt{unify}_E$ maps each $b$ to the smallest literal equivalent to $b$ given $E$. We can compute $\texttt{unify}_E$ in almost linear (amortized) time using a variation of the union-find algorithm (Tarjan, 1975).

**Example 6.** *For the equi-formula $E$ and substitution $\theta$ from Example 5 we have that $\texttt{unify}_E = \theta$.*

The following proposition provides the foundation for equi-propagation based Boolean simplification. It allows us to apply equational information to simplify a given formula. In particular, if $E$ is an equi-formula about literals occurring in $\varphi$ then $\texttt{unify}_E(\varphi)$ is smaller than $\varphi$ in that it contains fewer variables.

**Proposition 1.** *Let $\varphi$ be a Boolean formula and $E \in \mathcal{E}$ be a satisfiable equi-formula. Then,*

*a.* $\varphi \wedge E \;\leftrightarrow\; \texttt{unify}_E(\varphi) \wedge E$*;*

*b.* $\varphi \wedge E$ *is satisfiable if and only if* $\texttt{unify}_E(\varphi)$ *is satisfiable; and*

*c. if* $\sigma$ *is a satisfying assignment for* $\texttt{unify}_E(\varphi)$ *then* $\sigma \circ \texttt{unify}_E$ *is a satisfying assignment for* $\varphi \wedge E$*.*

*Proof.* (a) Let $\theta = \texttt{unify}_E$ and assume that $\sigma$ is a satisfying assignment of $E$, then we can view $\sigma$ as a substitution, and as a unifier of $E$. Hence, since $\theta$ is a most general unifier, there exists a substitution $\gamma$ such that $\sigma = \gamma \circ \theta$. Clearly $\gamma(b) = \sigma(b)$ for all variables $b$ in the range of $\theta$. Hence, $\sigma$ and $\gamma$ agree on all variables in $\theta(\varphi)$ which implies that $\sigma(\theta(\varphi)) = \gamma(\theta(\varphi))$ meaning that $\sigma(\theta(\varphi)) = \sigma(\varphi)$. So, $\sigma$ is a satisfying assignment of $\theta(\varphi) \wedge E$ if and only if $\sigma$ is a satisfying assignment of $\varphi \wedge E$. (b) The ($\rightarrow$) direction follows from (a) and the ($\leftarrow$) direction from (c). (c) Assume $\sigma$ is a satisfying assignment of $\texttt{unify}_E(\varphi)$. Clearly $\sigma \circ \texttt{unify}_E$ satisfies $\varphi$ by construction. Also $\sigma \circ \texttt{unify}_E$ satisfies $E$ since $\texttt{unify}_E(E)$ is trivial. Hence $\sigma \circ \texttt{unify}_E$ is a satisfying assignment of $\varphi \wedge E$. □

#### 3.1.3 The Equi-propagation Process

The equi-propagation process presented now is a central theme in this paper: Let $\Phi = \varphi_1 \wedge \cdots \wedge \varphi_n$ be a partitioning of a Boolean formula to $n$ portions, let $\mu_{\varphi_1}, \ldots, \mu_{\varphi_n}$ be corresponding equi-propagators, and take initial $E = \emptyset$. Satisfiability of $\Phi$ can be determined as follows:

1. So long as possible, select $\varphi_i$ such that $\mu_{\varphi_i}(E) \supsetneq E$ and update $E = \mu_{\varphi_i}(E)$.
2. Finally, when the equi-propagators apply no more, check if $\texttt{unify}_E(\Phi)$ is satisfiable.
3. If $\eta$ is a satisfying assignment for $\texttt{unify}_E(\Phi)$ then $\texttt{unify}_E \circ \eta$ is a satisfying assignment for $\Phi$.

We typically apply this equi-propagation theme to the Boolean representation $\Phi = \varphi_1 \wedge \cdots \wedge \varphi_n$ of a constraint model $C = C_1 \wedge \cdots \wedge C_n$ where $\varphi_i = [\![C_i]\!]$. Here we require that each $C_i$ is a "small" conjunction of constraints. Typically, the integer variables referred to in each $C_i$ are also declared in $C_i$ (sometimes this requires duplicating the variable declarations). For an individual constraint $c$ we denote by $c^+$ the conjunction of constraints including $c$ and the declarations for integer variables it refers to. The specifics of these declarations will be clear from the context.

**Example 7.** *Let $C$ be the following constraint model:*

$$C = \left( \begin{array}{l} \texttt{new_int(X,1,3)} \;\wedge\; \texttt{new_int(Y,1,3)} \;\wedge\; \texttt{new_int(Z,1,3)} \;\wedge \\ \texttt{int_plus(X,Y,3)} \;\wedge\; \texttt{int_plus(Y,Z,4)} \;\wedge\; \texttt{int_neq(Y,Z)} \end{array} \right)$$

*We have*

1. $\texttt{int_plus}^+(\texttt{X},\texttt{Y},3) = \texttt{int_plus}(\texttt{X},\texttt{Y},3) \ \wedge\ \texttt{new_int}(\texttt{X},1,3)\ \wedge\ \texttt{new_int}(\texttt{Y},1,3)$,

2. $\texttt{int_plus}^+(\texttt{Y},\texttt{Z},4) = \texttt{int_plus}(\texttt{Y},\texttt{Z},4) \ \wedge\ \texttt{new_int}(\texttt{Y},1,3)\ \wedge\ \texttt{new_int}(\texttt{Z},1,3)$,

3. $\texttt{int_neq}^+(\texttt{Y},\texttt{Z}) = \texttt{int_neq}(\texttt{Y},\texttt{Z}) \ \wedge \texttt{new_int}(\texttt{Y},1,3)\ \wedge\ \texttt{new_int}(\texttt{Z},1,3)$.

*As a basis for equi-propagation we take $\Phi = \varphi_1 \wedge \varphi_2 \wedge \varphi_3$ where $\varphi_1 = [\![\texttt{int_plus}^+(\texttt{X},\texttt{Y},3)]\!]$, $\varphi_2 = [\![\texttt{int_plus}^+(\texttt{Y},\texttt{Z},4)]\!]$, and $\varphi_3 = [\![\texttt{int_neq}^+(\texttt{Y},\texttt{Z})]\!]$. Denoting $X = [1, x_2, x_3]$, $Y = [1, y_2, y_3]$, and $Z = [1, z_2, z_3]$ and applying corresponding complete equi-propagators and starting with $E_0 = \emptyset$ we have:*

1. $E_1 = \hat{\mu}_{\varphi_1}(E_0) = E_0 \cup \{x_3 = 0, y_3 = 0, x_2 = \neg y_2\}$;

2. $E_2 = \hat{\mu}_{\varphi_2}(E_1) = E_1 \cup \{z_2 = 1, y_2 = \neg z_3\}$;

3. $E_3 = \hat{\mu}_{\varphi_3}(E_2) = E_2 \cup \{y_2 = 0\}$.

*At this point equi-propagation applies no more, and $\textbf{\textit{unify}}_{E_3} = \{x_2 \mapsto 1, x_3 \mapsto 0, y_2 \mapsto 0, y_3 \mapsto 0, z_2 \mapsto 1, z_3 \mapsto 1\}$ . Now, $\textbf{\textit{unify}}_{E_3}(\Phi)$ is a tautology (all of the Boolean variables are determent by equi-propagation).*

The following theorem clarifies that the order in which equi-propagators are applied in the equi-propagation process does not influence the final result.

**Theorem 8.** *The equi-propagation process is confluent.*

*Proof.* Let $\Phi = \varphi_1 \wedge \cdots \wedge \varphi_n$ be a Boolean formula and $\mu_{\varphi_1}, \ldots, \mu_{\varphi_n}$ corresponding equi-propagators. Let $E_1 = \mu_{\varphi_{i_r}}(\mu_{\varphi_{i_{r-1}}}(\ldots \mu_{\varphi_{i_1}}(\emptyset)\ldots))$ and $E_2 = \mu_{\varphi_{j_s}}(\mu_{\varphi_{j_{s-1}}}(\ldots \mu_{\varphi_{j_1}}(\emptyset)\ldots))$ be two different applications of the equi-propagation process. So by construction, for each of the given equi-propagators, we have a property ($\star$): $\mu_{\varphi_i}(E_1) = E_1$ and $\mu_{\varphi_i}(E_2) = E_2$.

Now assume, in contradiction, that $E_1 \neq E_2$. Then w.l.o.g. there exists $e \in E_2$ where $e$ *not* $\in E_1$ (swap the roles of $E_1$ and $E_2$ if $E_2 \subset E_1$). $E_1 \subsetneq E_2$. Let us focus on the first step in the equi-propagation process leading to $E_2$ that introduced the equation $e \in E_2$ not introduced to $E_1$: So, there exists an $\ell < s$ such that $E = \mu_{\varphi_{j_\ell}}(\mu_{\varphi_{j_{\ell-1}}}(\ldots \mu_{\varphi_{j_1}}(\emptyset)\ldots)) \subseteq E_1$ and $e \in \mu_{\varphi_{\ell+1}}(E)$ but $e \notin E_1$. But, if $E \subseteq E_1$, then by the monotonicity of $\mu_{\varphi_{\ell+1}}$, we have that $\mu_{\varphi_{\ell+1}}(E) \subseteq \mu_{\varphi_{\ell+1}}(E_1)$ and hence $e \in \mu_{\varphi_{\ell+1}}(E_1)$ in contradiction to the construction with property ($\star$). ☐

The following proposition provides an alternative, more efficient to implement, definition for complete equi-propagation.

**Proposition 2.** *Let $\varphi$ be a Boolean formula and $\hat{\mu}_\varphi$ a complete equi-propagator for $\varphi$. Define for $E \in \mathcal{E}$,*

$$\bar{\mu}_\varphi(E) = E \cup \left\{ \ e \in \mathcal{L}_{0,1}^{eq} \ \middle|\ \textbf{\textit{unify}}_E(\varphi) \models e \ \right\}$$

*Then, $\hat{\mu}_\varphi(E) = \bar{\mu}_\varphi(E)$. That is, $\bar{\mu}_\varphi$ implements a complete equi-propagator for $\varphi$.*

*Proof.* For the first direction, ($\Rightarrow$): By definition, we have that $\hat{\mu}_\varphi(E) \rightarrow E$. We also have $\hat{\mu}_\varphi(E) \rightarrow \left\{\, e \,\middle|\, \mathtt{unify}_E(\varphi) \models e \,\right\}$ because by Proposition 1(a) $\varphi \wedge E \models \mathtt{unify}_E(\varphi)$. So, $\hat{\mu}_\varphi(E) \rightarrow \bar{\mu}_\varphi(E)$. For the other direction, ($\Leftarrow$): Let $e \in \bar{\mu}_\varphi(E)$. If $e \in E$ then the proof is straightforward. Otherwise, let $\mathtt{unify}_E(\varphi) \models e$ and assume in contrary that $e \notin \hat{\mu}_\varphi(E)$, or in other words that $\varphi \wedge E \not\models e$. This means that there exists an assignment $\sigma$ that satisfies $\varphi \wedge E$ but does not satisfy $e$. By Lemma 1(a), $\sigma$ also satisfies $\mathtt{unify}_E(\varphi) \wedge E$ and in particular $\sigma$ satisfies $\mathtt{unify}_E(\varphi)$. From our assumption that $\mathtt{unify}_E(\varphi) \models e$ we now have that $\sigma$ satisfies $e$. Contradiction. □

Computing $\bar{\mu}_\varphi$ is considerably more efficient than $\hat{\mu}_\varphi$ since we can simply examine the formula $\varphi$ after the application of $\mathtt{unify}_E$ to determine new Boolean equality consequences.

Finally we comment: Our intention is that the equi-propagation process be applied not only to make a SAT instance smaller but also to obtain an easier to solve representation. However, decreasing the size of the CNF is not the main objective. In fact, often we explicitly introduce redundancies to improve a SAT encoding. For example, consider an "if-then-else" construct, `x↔ITE(s,t,f)`, where propositional variable: `s` indicates the "selector", `t` indicates the "true branch", `f` indicates the "false branch", and `x` indicates the result. The corresponding CNF is $\{\{\neg s, \neg t, x\}, \{\neg s, t, \neg x\}, \{s, \neg f, x\}, \{s, f, \neg x\}\}$. Eén and Sörensson (2006) propose to add redundant clauses, $\{\neg t, \neg f, x\}$ and $\{t, f, \neg x\}$. They comment that this improves the encoding and they observe that redundant clauses are often introduced to achieve arc-consistency in the SAT encoding. We show that given a clausal encoding of some formula $\Phi$, application of equi-propagation can only strengthen unit propagation.

**Theorem 9.** *Let $C$ be a set of clauses, and suppose $C \models E$ where $E$ is an equi-formula. Then unit propagation on $\mathtt{unify}_E(C)$ is at least as strong as unit propagation on $C$.*

*Proof.* Unit propagation on $C$ starting from assignment $A_0$ repeatedly chooses a clause $c \cup \{l\} \in C$ where $\{\neg l' \mid l' \in c\} \subseteq A_i$ and sets $A_{i+1} := A_i \cup \{l\}$. Unit propagation terminates with $A_k$ when no such clauses occur. Note that failure is detected when $A_k$ contains both a literal and its negation.

We show that using a order of unit propagation on $\mathtt{unify}_E(C)$ determined by that which occurs on $C$ starting from assignment $B_0 = \mathtt{unify}_E(A_0)$ we always obtain an assignment $B_i$ where $B_i \supseteq \mathtt{unify}_E(A_i)$. The proof is by induction on the unit propagation steps in $C$. The base case holds by construction.

Assume $c \cup \{l\} \in C$ where $\{\neg l' \mid l' \in c\} \subseteq A_i$. Then by induction $B_i \supseteq \mathtt{unify}_E(A_i) \supseteq \{\mathtt{unify}_E(\neg l') \mid l' \in c\}$. Either $\mathtt{unify}_E(l) \in B_i$ in which case we set $B_{i+1} = B_i$ and the induction holds. Or $\mathtt{unify}_E(l) \notin B_i$. Now since $c \cup \{l\} \in C$ we have that $\{\mathtt{unify}_E(l') \mid l' \in c\} \cup \{\mathtt{unify}_E(l)\} \subseteq \mathtt{unify}_E(C)$. Hence by unit propagation on $\mathtt{unify}_E(C)$ and $B_i$ we obtain $B_{i+1} := B_i \cup \{\mathtt{unify}_E(l)\}$. Hence the induction holds.

Given that unit propagation reaches a unique fixpoint then any unit propagation order on $\mathtt{unify}_E(A_0)$ will end up with an assignment $B$ where $B \supseteq B_k \supseteq \mathtt{unify}(A_k)$ □

### 3.2 The Power of Equi-propagation

To illustrate the impact of equi-propagation we come back to the Kakuro example from Section 2.2 (recall Figure 2). In fact solving such puzzles via SAT encodings is quite easy, with and without equi-propagation. So the example should only be viewed as illustrating

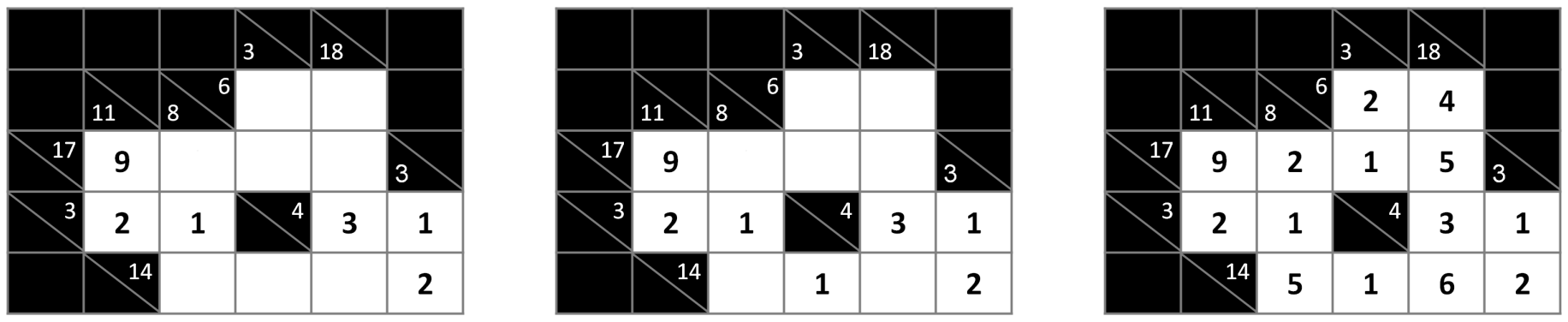

a. $\Phi_1$ b. $\Phi_2$ c. $\Phi_3$

Figure 4: Applying complete equi-propagation to a Kakuro Instance using different models

the impact of equi-propagation on the size of the encoding. We compare 3 different models of the problem, which each give different equi-propagation.

We consider, as a baseline for this discussion, the following Boolean representation derived from a constraint model where the declarations which are not specified explicitly are of the form $\texttt{new_int}(I, 1, h)$ where $h$ is the smallest hint for a block that includes $I$ or the number 9 if that is smaller.

$$\Phi_1 = \bigwedge_{\substack{\{I_1,\dots,I_k\} \in \mathit{Blocks} \\ 1 \leq i < j \leq k}} [\![\texttt{int_neq}^+(I_i, I_j)]\!] \;\wedge \bigwedge_{B \,\in\, \mathit{Blocks}} [\![\texttt{int_array_sum}^+(B, \mathit{hint}(B))]\!]$$

Notice that there is one "int_neq" conjunct for each pair of white cells in the same block, and one "int_array_sum" conjunct for each block. Applying the equi-propagation process to $\Phi_1$ with complete equi-propagators determines six integer values as depicted in Figure 4(a).

Figure 4(b) illustrates the impact of applying the equi-propagation process where the equi-propagators are for allDiff constraints instead of for the individual int_neq constraints. This determines seven integer variables and is formalized taking the following Boolean representation of the constraint model (and introducing an equi-propagator for each conjunct).

$$\Phi_2 = \bigwedge_{B \,\in\, \mathit{Blocks}} [\![\texttt{allDiff}^+(B)]\!] \quad \wedge \bigwedge_{B \,\in\, \mathit{Blocks}} [\![\texttt{int_array_sum}^+(B, \mathit{hint}(B))]\!]$$

Figure 4(c) illustrates the impact of applying the equi-propagation process where the equi-propagators are for pairs, each consisting of an allDiff constraint together with its corresponding sum constraint. This form of equi-propagation is most powerful. It fixes integer values for all of the white cells (in this example). We stress that equi-propagation reasons only about equalities between Boolean literals and constants. Here we take the model as:

$$\Phi_3 = \bigwedge_{B \,\in\, \mathit{Blocks}} \left([\![\texttt{allDiff}^+(B)]\!] \;\wedge\; [\![\texttt{int_array_sum}^+(B, \mathit{hint}(B))]\!]\right)$$

To further demonstrate the impact of equi-propagation, Table 1 provides data for 15 additional instances,[1] categorized as: "easy", "medium" and "hard". The first two columns in the table indicate the instance category and ID. From the five columns headed "Integer

1. Instances available from http://4c.ucc.ie/~hcambaza/page1/page7/page7.html (generated by Helmut Simonis).

| | | Integer Variables | | | | | Boolean Variables | | | | |
|---|---|---|---|---|---|---|---|---|---|---|---|
| | ID | init | $\Phi_1$ | $\Phi_2$ | $\Phi_3$ | BEE | init | $\Phi_1$ | $\Phi_2$ | $\Phi_3$ | BEE |
| easy | 168 | 484 | 439 | 280 | 0 | 385 | 3872 | 1440 | 843 | 0 | 1170 |
| | 169 | 467 | 456 | 440 | 0 | 440 | 3736 | 1823 | 1682 | 0 | 1692 |
| | 170 | 494 | 485 | 469 | 0 | 469 | 3952 | 1961 | 1798 | 0 | 1805 |
| | 171 | 490 | 406 | 393 | 0 | 422 | 3920 | 1280 | 1148 | 0 | 1341 |
| | 172 | 506 | 495 | 484 | 0 | 492 | 4048 | 1676 | 1573 | 0 | 1634 |
| medium | 188 | 476 | 461 | 455 | 0 | 461 | 3808 | 1939 | 1915 | 0 | 1934 |
| | 189 | 472 | 437 | 425 | 62 | 449 | 3776 | 2017 | 1911 | 81 | 1976 |
| | 190 | 492 | 481 | 480 | 0 | 480 | 3936 | 1998 | 1920 | 0 | 1936 |
| | 191 | 478 | 452 | 448 | 161 | 448 | 3824 | 1864 | 1821 | 197 | 1828 |
| | 192 | 499 | 481 | 478 | 136 | 478 | 3992 | 2455 | 2417 | 214 | 2420 |
| hard | 183 | 490 | 365 | 345 | 0 | 371 | 3920 | 1151 | 1059 | 0 | 1168 |
| | 184 | 506 | 489 | 484 | 23 | 486 | 4048 | 1613 | 1495 | 21 | 1545 |
| | 185 | 482 | 482 | 455 | 206 | 467 | 3856 | 2181 | 2111 | 220 | 2144 |
| | 186 | 472 | 466 | 454 | 0 | 466 | 3776 | 2115 | 2062 | 0 | 2086 |
| | 187 | 492 | 475 | 473 | 69 | 473 | 3936 | 1991 | 1959 | 48 | 1960 |
| | Average compilation time in sec. | | | | | | | 3.739 | 2.981 | 0.916 | 0.477 |

Table 1: Applying SAT-based complete equi-propagation on Kakuro encoding

Variables", the first four specify the number of unassigned white cells in the initial stage and after each of the three complete equi-propagation processes described above. From the five columns headed "Boolean variables", the first four indicate the corresponding information regarding the number of Boolean variables in the bit representations of the integers. So, the smaller the number in the table, the more variables have been removed due to equi-propagation. In particular, the $\Phi_3$ model completely solves 9 of the 15 instances. The two columns titled BEE show the corresponding information obtained using a weaker form of equi-propagation that is described in Section 4 below. The last row of the table indicates the average time it takes to perform equi-propagation (in seconds) using each of the three schemes, $\Phi_1, \Phi_2, \Phi_3$, and the weaker scheme titled BEE. We will come back to discuss this later after detailing how equi-propagation is performed. The results in the table indicate the clear benefit in performing equi-propagation based on coarser portions of the model.

### 3.3 Implementing Equi-propagators

To implement complete equi-propagators we need to infer Boolean equalities implied by a given Boolean formula, $\varphi$, and equi-formula, $E$. Based on Proposition 2, it is sufficient to test for the condition

$$\texttt{unify}_E(\varphi) \models (\ell_1 \leftrightarrow \ell_2) \tag{1}$$

We consider three techniques: using a SAT solver, using BDD's, and using ad-hoc rules applied to the Boolean representations of individual constraints.

It is straightforward to implement a complete equi-propagator using a SAT solver. To test Condition (1) we consider the formula $\psi = \varphi \wedge (\ell_1 \not\leftrightarrow \ell_2)$. If $\psi$ is not satisfiable, then Condition (1) holds. In this way, Condition (1) can be checked for all relevant equations involving variables from $\texttt{unify}_E(\varphi)$ (and constants 0,1). A major obstacle with this SAT-based approach is that testing for a single equivalence, $\ell_1 \leftrightarrow \ell_2$, is at least as hard as testing for the satisfiability of $\varphi$. In fact testing for unsatisfiability is typically more expensive. Hence the importance of our assumption that $\varphi$ is only a small fragment of the CNF of interest. In practice SAT-based equi-propagation is surprisingly fast. For illustration, in the last row of Table 1 the average times for SAT-based complete equi-propagation for the different models are indicated in the columns $\Phi_1$, $\Phi_2$, and $\Phi_3$. It is interesting to observe that the strongest technique, using $\Phi_3$, is the fastest. This is because there are fewer (but larger) conjuncts and hence fewer queries to the SAT solver.

We can implement a complete equi-propagator using binary decision diagrams (BDDs) as follows. We construct a BDD for formula $\varphi$ at the beginning of equi-propagation. When new equational information $E'$ is added to $E$ we "simplify" the BDD for $\varphi$ by conjoining the BDD with a BDD for $E'$ and then projecting out the variables that no longer appear in $\texttt{unify}_E(\varphi)$. Note that this "simplification" can increase the size of the BDD. In practice, rather than these two steps, we can use the "Restrict" operation of Coudert and Madre (1990) ("`bdd_simplify`" in Somenzi, 2009) to create the new BDD more efficiently.

Given the BDD for $\texttt{unify}_E(\varphi)$, we can explicitly test Condition (1) using a standard BDD containment test (e.g., "`bddLeq`" in Somenzi, 2009). Just as in the SAT-based approach, this test is performed for all relevant equations involving variables from $\texttt{unify}_E(\varphi)$ (and constants 0,1). Alternately we can use the method of Bagnara and Schachte (1998) (extended to extract literal equalities as opposed to just variable equalities) to extract all the fixed literals and equivalent literal consequences of the BDD.

**Example 10.** *Consider the BDD shown in Figure 5(a) which represents the formula:* $\varphi \equiv \textit{new_int}(A,0,3) \wedge \textit{new_int}(B,0,3) \wedge \textit{int_neq}(A,B)$. *Figure 5(b) depicts the The BDD for* $\textit{unify}_E(\varphi)$ *where* $E = \{B_1 = 1,\ B_2 = 1,\ B_3 = 0\ \}$. *Here it is easy to see that equi-propagation determines that* $A_2 = A_3$. *Let* $E' = E \cup \{A_2 = A_3\}$. *Then Figure 5(c) shows the simplified BDD for* $\textit{unify}_{E'}(\varphi)$.

A major obstacle with this BDD-based approach concerns the size of the formula $\texttt{unify}_E(\varphi)$. For some constraints, the corresponding BDD is guaranteed to be polynomial (in the size of the constraint). The following result holds for an arbitrary constraint $\varphi$, so it also holds for $\texttt{unify}_E(\varphi)$.

**Proposition 3.** *Let $c$ be a constraint about $k$ integer variables each represented with $n$ bits in the order encoding. Then, the number of nodes in the BDD representing $[\![c]\!]$ is bound by $O(n^k)$.*

*Proof.* (Sketch) There are only $n+1$ legitimate states for each $n$ bit unary variable, and the BDD cannot have more nodes than possible states. □

Constraints like `new_int`, `int_neq`, and `int_plus` involve at most 3 integer variables and hence their BDD-based complete equi-propagators are polynomially bounded. However, this is not the case for global constraints such as `allDiff` and `int_array_plus` where the arity is not fixed. Moreover, it is well known that the `allDiff` constraint does not have a polynomial sized BDD (Bessiere, Katsirelos, Narodytska, & Walsh, 2009).

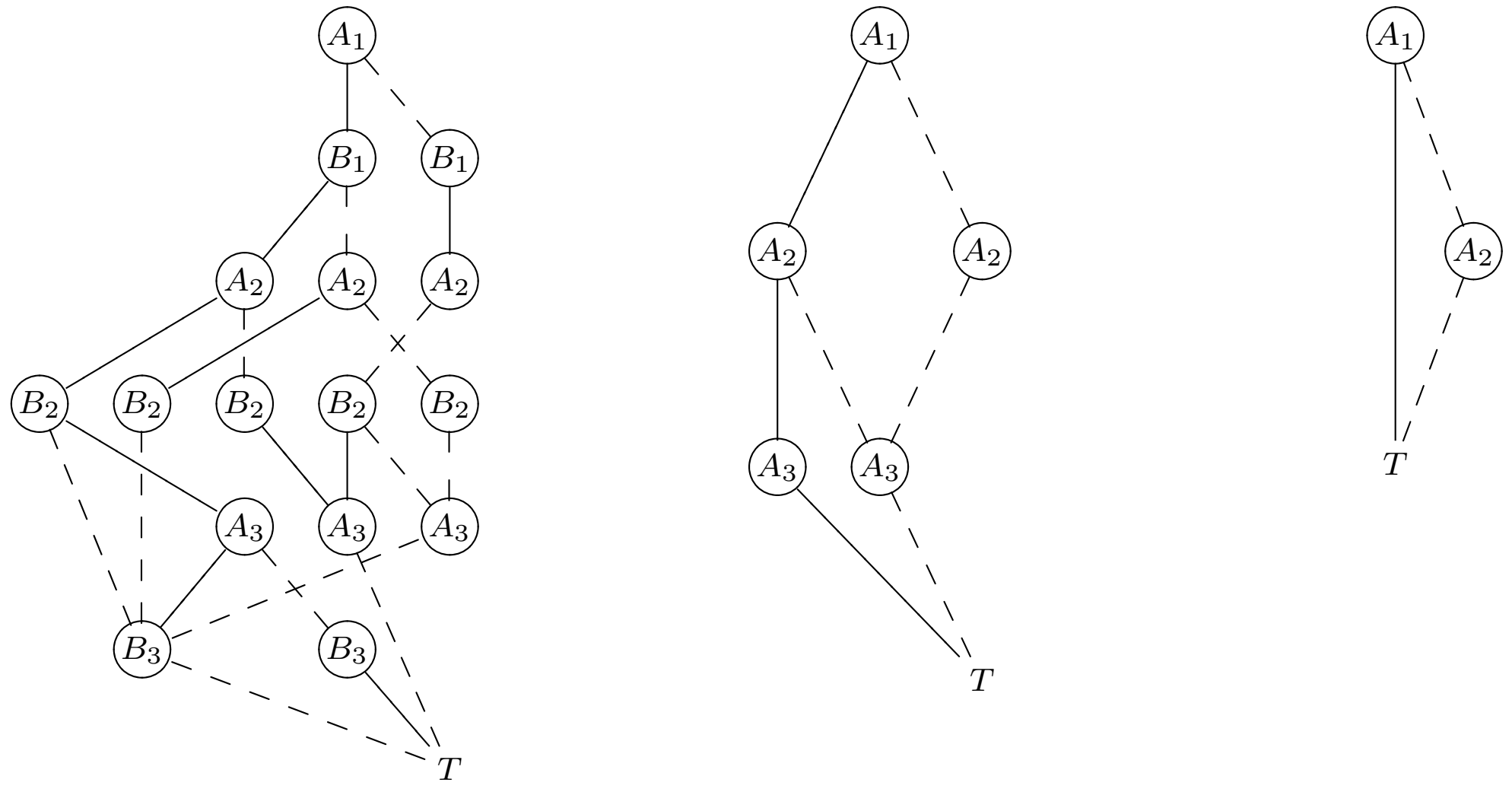

**(a)** BDD for $\texttt{int_neq}_u(A, B)$ **(b)** Simpl'd wrt $B$=[1, 1, 0] **(c)** Simpl'd wrt $A_2$=$A_3$

Figure 5: BDDs for (a) $\varphi \equiv \texttt{new_int}_3(A, [0, 3]) \wedge \texttt{new_int}_3(B, [0, 3]) \wedge \texttt{int_neq}(A, B)$ (b) $\texttt{unify}_E(\varphi)$ where $E = \{B_1{=}1, B_2{=}1, B_3{=}0\}$ and (c) $\texttt{unify}_{E'}(\varphi)$ where $E' = E \cup \{A_2{=}A_3\}$. Full (dashed) lines correspond to true (false) edges. Edges to the false node "F" are omitted for brevity.

Given the potential exponential run-time when performing SAT-based equi-propagation, and the potential exponential size of BDD-based equi-propagators, we consider a third approach where we implement equi-propagation by a collection of ad-hoc transition rules for each type of constraint. While this approach is not complete — there are equations implied by a constraint that are not detected — the implementation is fast, and works well in practice. This is the topic of the next section.

## 4. Ad-hoc Equi-Propagation

We consider a rule-based approach to define equi-propagators. The definition is given as a set of ad-hoc rules specified for each type of constraint. The novelty is that the approach is not based on CNF, as in previous works, but rather driven by the bit blasted constraints that are to be encoded to CNF. Our presentation focuses on the case where finite domain integers are represented in the order encoding. For an integer $X = [x_1, \dots, x_n]$, we often write: $X \geq i$ to denote the equation $x_i = 1$, $X < i$ to denote the equation $x_i = 0$, $X \neq i$ to denote the equation $x_i = x_{i+1}$, and $X = i$ to denote the pair of equations $x_i = 1, x_{i+1} = 0$. Moreover, to simplify notation when specifying the rules below, we view $X = [x_1, \dots, x_n]$ as a larger vector padded with sentinel cells such that all cells "to the left of" $x_1$ take value 1 and all cells "to the right of" $x_n$ take the value 0. Basically this facilitates the specification of the "end cases" in our formalism. We now consider each of the 5 constraints in the language fragment presented in Section 2.

| $c = \texttt{new_int}([x_1, \dots, x_n], 0, n)$ | |
|---|---|
| if in E | then add in $\mu_c$(E) |
| $x_i = 1$ | $x_1 = 1, \dots, x_{i-1} = 1$ |
| $x_i = 0$ | $x_{i+1} = 0, \dots, x_n = 0$ |

(a)

| $c = \texttt{int_neq}(X, Y)$ where $X = [x_1, \dots, x_n]$ and $Y = [y_1, \dots, y_n]$ | |
|---|---|
| if in E | then add in $\mu_c$(E) |
| $X = i$ | $Y \neq i$ |
| $x_i = y_{i+1}$, $y_i = x_{i+1}$ | $X \neq i$, $Y \neq i$ |
| $x_i = \neg y_{i+1}$, $y_i = \neg x_{i+1}$ | $X \neq i$, $Y \neq i$ |

(b)

Figure 6: Ad-hoc rules for (a) new_int and (b) int_neq

| $c = \texttt{allDiff}([Z_1, Z_2, Z_3, \dots, Z_n])$ | |
|---|---|
| if in E | then add in $\mu_c$(E) |
| $Z_1, Z_2 \in \{i, j\}$ | $Z_1 \neq Z_2$, $Z_k \neq i$<br>$Z_k \neq j$ $(k > 2)$ |

(a)

| $c = \texttt{int_plus}(X, Y, Z)$ where $X = [x_1, \dots, x_n]$, $Y = [y_1, \dots, y_m]$, and $Z = [z_1, \dots, z_{n+m}]$ | |
|---|---|
| if in E | then add in $\mu_c$(E) |
| $X \geq i$, $Y \geq j$ | $Z \geq i + j$ |
| $X < i$, $Y < j$ | $Z < i + j - 1$ |
| $Z \geq k$, $X < i$ | $Y \geq k - i$ |
| $Z < k$, $X \geq i$ | $Y < k - i$ |
| $X = i$ | $z_{i+1} = y_1, \dots, z_{i+m} = y_m$ |
| $Z = k$ | $x_1 = \neg y_k, \dots, x_k = \neg y_1$ |

(b)

Figure 7: Ad-hoc rules for (a) allDiff and (b) int_plus

**(1)** The two rules in Figure 6(a) derive from the monotonicity in the order encoding representation. These basically correspond to unit propagation, but at the constraint level.

**(2)** The first rule in Figure 6(b) considers cases when $X$ is a constant (the symmetric case can be handled by exchanging $X$ and $Y$). The other two rules capture templates that commonly arise in the equi-propagation process. To illustrate the justification of the third rule consider all possible truth values for the variables $x_i$ and $x_{i+1}$: (a) If $x_i = 0$ and $x_{i+1} = 1$ then both integers in the relation take the form $[\dots, 0, 1, \dots]$ violating their specification as `ordered`, so this is not possible. (b) If $x_i = 1$ and $x_{i+1} = 0$ then both numbers take the form $[1, \dots, 1, 0, \dots, 0]$ and are equal, violating the `neq` constraint. The only possible bindings for $x_i$ and $x_{i+1}$ are those where $x_i = x_{i+1}$.

**(3)** In Figure 7(a) we illustrate a single rule for the `allDiff` constraint which considers Hall sets of size 2. Here each $Z_i$ represents an integer in the order encoding and we focus on the case when $Z_1$ and $Z_2$ are restricted by the equations in $E$ to take only two possible values, $i$ or $j$. This can be expressed in $E$ because $[x_1, \dots, x_n] \in \{i, j\}$ (for $i < j$) means that $x_k = 1$ for $k < i$, $x_k = x_{k+1}$ for $i \leq k < j$, and $x_k = 0$ for $j < k \leq n$. $Z_1 \neq Z_2$ then means adding the single equation $x_i = \neg y_i$ (because $Z_1$ and $Z_2$ can take only two values). In addition to this rule, we apply the rules for $\texttt{int_neq}(Z_i, Z_j)$ for each pair of integers $Z_i$ and $Z_j$ in the constraint.

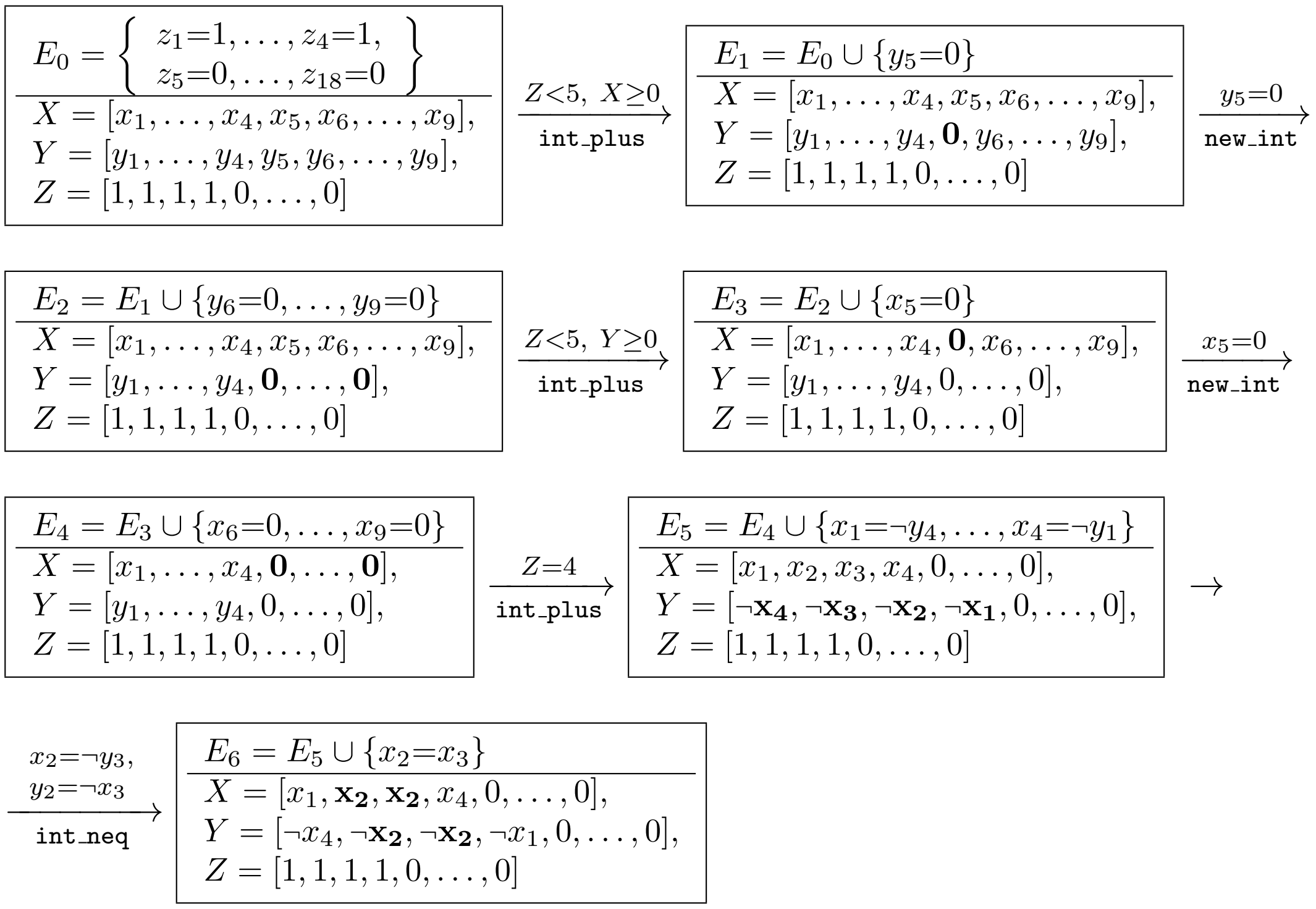

Figure 8: Ad-hoc equi-propagation described in Example 11

**(4)** The first four rules of Figure 7(b) capture the standard propagation behavior for interval arithmetics. The last two rules apply when one of the integers in the relation is a constant. There are symmetric cases when replacing the role of $X$ and $Y$.

**(5)** There are no special ad-hoc rules for equi-propagation of an `int_array_plus` constraint. These are simply viewed as a decomposition to a set of `int_plus` constraints. Then simplification is performed at that level using the rules for `int_plus`. The decomposition of `int_array_plus` is explained in Section 6.

**Example 11** (ad-hoc equi-propagation)**.** *Consider the following (partial) constraint model, from the context of the Kakuro example of Section 2.2, where we represent variables $X$, $Y$ and $Z$ as $X = [x_1, \ldots, x_9]$, $Y = [y_1, \ldots, y_9]$ and $Z = [z_1, \ldots, z_{18}]$ and assume some previous equi-propagation (on other constraints) has determined the current equi-formula $E_0$ to specify that integer variable $Z = 4$:*

$$C = \left( \begin{array}{l} \texttt{new_int(X,0,9)} \ \wedge \ \texttt{new_int(Y,0,9)} \ \wedge \ \texttt{new_int(Z,0,18)} \ \wedge \\ \texttt{int_plus(X,Y,Z)} \ \wedge \ \texttt{int_neq(X,Y)} \end{array} \right)$$

*Figure 8 illustrates, step-by-step, the equi-propagation process on $C$ using the ad-hoc rules defined above. Each step corresponds to the application of one of the above defined ad-hoc equi-propagation rules as indicated by the label on the transition. At each stage we illustrate the derived equations (top part) and their application (as a unifier) to the state variables $X, Y$ and $Z$ (lower part).*

| $c = \texttt{ordered}([x_1, \ldots, x_n])$ (`new_int`) | |
|---|---|
| if | then replace with |
| $n \leq 1$ | `true` |
| $x_1 = 1$ | $\texttt{ordered}([\cancel{1}, x_2 \ldots, x_n])$ |
| $x_n = 0$ | $\texttt{ordered}([x_1, \ldots, x_{n-1}, \cancel{0}])$ |
| $x_i = x_{i+1}$ | $\texttt{ordered}([x_1, \ldots, x_i, \cancel{x_{i+1}}, \ldots, x_n])$ |

Figure 9: Simplification rules for `new_int` (crossed out elements have been removed).

To summarize, let us come back to Table 1. The numbers presented in the two columns headed "BEE" specify the number of variables remaining after application of ad-hoc equi-propagation. We also observe that our definition of ad-hoc equi-propagation is trivially monotonic.

## 5. Constraint Model Partial Evaluation

Partial evaluation, together with equi-propagation, is the second important component in our approach to compile constraint models to CNF. Partial evaluation is about simplifying a given constraint model in view of information that becomes available due to equi-propagation. Typically, in the constraint simplification process, we apply alternating steps of equi-propagation and partial evaluation. Examples of partial evaluation include constant elimination and removing constraints which are tautologies. In this section we detail the partial evaluation rules that apply for the five constraint types defined in the language fragment presented in Section 2.

**(1)** A $\texttt{new_int}(\texttt{I}, \texttt{c}_1, \texttt{c}_2)$ constraint specifies that an integer $\texttt{I} = [\texttt{x}_1, \ldots, \texttt{x}_n]$ is represented in the order encoding and in particular that the corresponding bit sequence is sorted (not increasing). We denote this as $\texttt{ordered}([\texttt{x}_1, \ldots, \texttt{x}_n])$. Partial evaluation focuses on this aspect of the constraint and ignores the bounds $\texttt{c}_1$, $\texttt{c}_2$ specified in the constraint. The table in Figure 9 specifies four simplification rules that apply. The first rule identifies tautologies, the second and third rules remove leading ones and trailing zeros, and the fourth removes (one of two) equated bits. In this figure, and in the subsequent, a crossed out element in a sequence, indicates that it has been removed from the sequence.

**(2)** The simplification rules for a `int_neq` constraint shown in Figure 10(a) are symmetric when exchanging the role of $X$ and $Y$. The first two rules identify tautologies. The third rule is about $X$ and $Y$ which have an equal bit at position $i$. The corresponding bits can be removed from the representation of $X$ and $Y$, resulting in a shorter list of bits in their representations. The last two rules are about removing leading ones and trailing zeroes and are illustrated by the following example.

**Example 12.** *Figure 10(b) shows two steps of partial evaluation, for a* `int_neq` *constraint, first removing leading ones, then removing trailing zeroes.*

| $c = \texttt{int_neq}(X, Y)$ where $X = [x_1, \dots, x_n]$ and $Y = [y_1, \dots, y_n]$ | |
|---|---|
| if | then replace with |
| $X = i$, $Y \neq i$ | `true` |
| $x_i = \neg y_i$ | `true` |
| $x_i = y_i$ | $\texttt{int_neq}([x_1, \dots, \cancel{x_i}, \dots, x_n], [y_1, \dots, \cancel{y_i}, \dots, y_n])$ |
| $X \geq i \geq 2$ | $\texttt{int_neq}([1, x_{i+1}, \dots, x_n], [y_i, y_{i+1}, \dots, y_n])$ |
| $X \leq i$ | $\texttt{int_neq}([x_1, \dots, x_i, 0], [y_1, \dots, y_i, y_{i+1}])$ |

(a)

$$\begin{bmatrix} \texttt{int_neq}( \\ \quad [\mathtt{x}_1, \dots, \mathtt{x}_4, 0, 0, 0], \\ \quad [1, 1, 1, \mathtt{y}_4, \dots, \mathtt{y}_7]) \end{bmatrix} \xrightarrow[\texttt{int_neq}]{P.E} \begin{bmatrix} \texttt{int_neq}( \\ \quad [\mathtt{x}_3, \mathtt{x}_4, 0, 0, 0], \\ \quad [1, \mathtt{y}_4, \dots, \mathtt{y}_7]) \end{bmatrix} \xrightarrow[\texttt{int_neq}]{P.E} \begin{bmatrix} \texttt{int_neq}( \\ \quad [\mathtt{x}_3, \mathtt{x}_4, 0], \\ \quad [1, \mathtt{y}_4, \mathtt{y}_5]) \end{bmatrix}$$

(b)

Figure 10: (a) Simplification rules for `int_neq` and (b) an example of their application.

| $c = \texttt{allDiff}([Z_1, \dots, Z_n])$ where $Z_i = [z_{i,1}, \dots, z_{i,m}]$ $(1 \leq i \leq n)$ | |
|---|---|
| if | then replace with |
| $n \leq 1$ | `true` |
| $\bigwedge_{k>1} \left( \begin{array}{c} dom(Z_1) \cap \\ dom(Z_k) = \emptyset \end{array} \right)$ | $\texttt{allDiff}([Z_2, \dots, Z_n])$ |
| $\lvert \bigcup_{i \in \{1,2\}} dom(Z_i) \rvert = 2$ | $\texttt{allDiff}([Z_3, \dots, Z_n])$ |
| $\bigwedge_k Z_k \neq i$ | $\texttt{allDiff}([z_{1,1}, \dots, \cancel{z_{1,i+1}}, \dots, z_{1,m}] \dots [z_{n,1}, \dots, \cancel{z_{n,i+1}}, \dots, z_{n,m}])$ |

Figure 11: Simplification rules for `allDiff`

**(3)** Four rules for simplifying `allDiff` constraints are illustrated in Figure 11. The first, is about detecting tautologies. The second, identifies cases when one of the integers in the constraint (assume $Z_1$) has a domain disjoint from all of the others. This rule also captures the case when $Z_1$ is a constant. The third rule removes a Hall set of size 2 (assume $\{Z_1, Z_2\}$) from the constraint. Note that the corresponding equi-propagation rule detects that the values of $Z_3, \dots, Z_n$ are different from the values of $\{Z_1, Z_2\}$ and then the next fourth rule applies. The fourth rule is for the case when none of the integers in the constraint can take a certain value $i$. This rule also captures the case when all of the numbers have leading ones or trailing zeroes. The last two rules are illustrated in Example 14.

**(4 & 5)** The simplification rules shown in Figure 12 are symmetric when exchanging the role of $X$ and $Y$. The first two apply where (at least) one of $X$, $Y$ and $Z$ is a constant. Because we have already applied equi-propagation to the constraint, it is a tautology. See Example 13. The last two rules apply to remove leading ones and trailing zeroes. The

| $c = \texttt{int_plus}(X,Y,Z)$ where $X = [x_1,\dots,x_n]$, $Y = [y_1,\dots,y_m]$, and $Z = [z_1,\dots,z_{n+m}]$ | |
|---|---|
| if | then replace with |
| $X = i$ | `true` |
| $Z = k$ | `true` |
| $X \geq i$, $Z \geq i$ | $\texttt{int_plus}([x_{i+1},\dots,x_n], Y, [z_{i+1},\dots,z_{n+m}])$ |
| $X \leq i$, $Z \leq i+m$ | $\texttt{int_plus}([x_1,\dots,x_i], Y, [z_1,\dots,z_{i+m}])$ |

Figure 12: Simplification rules for `int_plus`.

(a) $\texttt{int_plus}(\texttt{I}_1, \texttt{I}_2, \texttt{K})$

(b) $\texttt{allDiff}([\texttt{I}_1, \texttt{I}_2, \texttt{I}_3, \texttt{I}_4, \texttt{I}_5, \texttt{I}_6, \texttt{I}_7, \texttt{I}_8])$

(c) $\texttt{int_array_plus}([\texttt{I}_2, \texttt{I}_3, \texttt{I}_4, \texttt{I}_5], \texttt{K})$

Figure 13: Constraint Model for Examples 13–15

simplification rules of an `int_array_plus` constraint are straightforward generalizations of the ones for `int_plus`. See Example 15.

To summarise the rule based approach to apply equi-propagation and partial evaluation we present the following sequence of three examples which focus on the simplification of the three constraints given as Figure 13 where the integer variables $\texttt{I}_1, \dots, \texttt{I}_8$ are defined in the range between 1 and 8 and where $K = 14$.

**Example 13.** *Consider equi-propagation of constraint (a) from Figure 13 where $E_0$ specifies that $K = 14$:*

$$\begin{array}{|l|}\hline \texttt{E}_0 = \left\{ \begin{array}{l} k_1{=}1,\dots,k_{14}{=}1 \\ k_{15}=0, k_{16}=0 \end{array} \right\} \\ \hline \texttt{I}_1 = [1, \texttt{i}_{1,2}, \dots, \texttt{i}_{1,8}], \\ \texttt{I}_2 = [1, \texttt{i}_{2,2}, \dots, \texttt{i}_{2,8}], \\ \texttt{K} = [\underbrace{1,1,\dots,1}_{14}, 0, 0] \\ \hline \end{array} \xrightarrow[\texttt{int_plus}]{\texttt{K=14}} \begin{array}{|l|}\hline \texttt{E}_1 = \texttt{E}_0 \cup \left\{ \begin{array}{l} i_{1,2}{=}1,\dots,i_{1,6}{=}1, \\ i_{2,2}{=}1,\dots,i_{2,6}{=}1, \\ i_{1,7}{=}\neg i_{2,8}, i_{1,8}{=}\neg i_{2,7} \end{array} \right\} \\ \hline \texttt{I}_1 = [1,1,1,1,1,1,\texttt{i}_{1,7},\texttt{i}_{1,8}], \\ \texttt{I}_2 = [1,1,1,1,1,1,\neg\texttt{i}_{1,8},\neg\texttt{i}_{1,7}], \\ \texttt{K} = [\underbrace{1,1,\dots,1}_{14}, 0, 0] \\ \hline \end{array}$$

*Given $E_1$, the constraint is a tautology and removed by partial evaluation:*

$$\left[ \begin{array}{l} \texttt{int_plus}( \\ \quad [1,1,1,1,1,1,\texttt{i}_{1,7},\texttt{i}_{1,8}], \\ \quad [1,1,1,1,1,1,\neg\texttt{i}_{1,8},\neg\texttt{i}_{1,7}],\ 14) \end{array} \right] \xrightarrow[\texttt{int_plus}]{P.E} [\quad]$$

**Example 14.** *Consider equi-propagation of constraint (b) from Figure 13 given $E_1$ from Example 13:*

$$\begin{array}{|l|}\hline \texttt{E}_1 \\ \hline \texttt{I}_1 = [1,1,1,1,1,1,\texttt{i}_{1,7},\texttt{i}_{1,8}], \\ \texttt{I}_2 = [1,1,1,1,1,1,\neg\texttt{i}_{1,8},\neg\texttt{i}_{1,7}] \\ \hline \end{array} \xrightarrow[\textit{int_neq}]{\begin{array}{c}\texttt{i}_{1,7}{=}\neg\texttt{i}_{2,8}, \\ \texttt{i}_{2,7}{=}\neg\texttt{i}_{1,8}\end{array}} \begin{array}{|l|}\hline \texttt{E}_2 = \texttt{E}_1 \cup \{\texttt{i}_{1,7}{=}\texttt{i}_{1,8}\} \\ \hline \texttt{I}_1 = [1,1,1,1,1,1,\texttt{i}_{1,7},\texttt{i}_{1,7}], \\ \texttt{I}_2 = [1,1,1,1,1,1,\neg\texttt{i}_{1,7},\neg\texttt{i}_{1,7}] \\ \hline \end{array}$$

*Given $E_2$, the equi-propagation rule for* $\mathit{allDiff}$ *detects that* $\{\mathtt{I}_1, \mathtt{I}_2\}$ *is a Hall set (where the two variables take values 6 and 8). and adds to $E_2$ the set of equations, $E'$, that specify that* $\mathtt{I}_3, \mathtt{I}_4, \mathtt{I}_5, \mathtt{I}_6, \mathtt{I}_7, \mathtt{I}_8 \neq 6, 8$. *The result is $E_3 = E_2 \cup E'$ and the result of this step gives the following bindings (where the impact of $E'$ is underlined):*

$$
\begin{array}{ll}
\mathtt{I}_1 = [1,1,1,1,1,1,\mathtt{i}_{1,7},\mathtt{i}_{1,7}] & \mathtt{I}_5 = [1,\mathtt{i}_{5,2},\mathtt{i}_{5,3},\mathtt{i}_{5,4},\mathtt{i}_{5,5},\underline{\mathtt{i}_{5,7},\mathtt{i}_{5,7},0}] \\
\mathtt{I}_2 = [1,1,1,1,1,1,\neg\mathtt{i}_{1,7},\neg\mathtt{i}_{1,7}] & \mathtt{I}_6 = [1,\mathtt{i}_{6,2},\mathtt{i}_{6,3},\mathtt{i}_{6,4},\mathtt{i}_{6,5},\underline{\mathtt{i}_{6,7},\mathtt{i}_{6,7},0}] \\
\mathtt{I}_3 = [1,\mathtt{i}_{3,2},\mathtt{i}_{3,3},\mathtt{i}_{3,4},\mathtt{i}_{3,5},\underline{\mathtt{i}_{3,7},\mathtt{i}_{3,7},0}] & \mathtt{I}_7 = [1,\mathtt{i}_{7,2},\mathtt{i}_{7,3},\mathtt{i}_{7,4},\mathtt{i}_{7,5},\underline{\mathtt{i}_{7,7},\mathtt{i}_{7,7},0}] \\
\mathtt{I}_4 = [1,\mathtt{i}_{4,2},\mathtt{i}_{4,3},\mathtt{i}_{4,4},\mathtt{i}_{4,5},\underline{\mathtt{i}_{4,7},\mathtt{i}_{4,7},0}] & \mathtt{I}_8 = [1,\mathtt{i}_{8,2},\mathtt{i}_{8,3},\mathtt{i}_{8,4},\mathtt{i}_{8,5},\underline{\mathtt{i}_{8,7},\mathtt{i}_{8,7},0}]
\end{array}
$$

*Given $E_3$, partial evaluation of the constraint first removes the Hall set:*

$$[\mathit{allDiff}([\mathtt{I}_1,\mathtt{I}_2,\mathtt{I}_3,\mathtt{I}_4,\mathtt{I}_5,\mathtt{I}_6,\mathtt{I}_7,\mathtt{I}_8])] \xrightarrow[\mathit{allDiff}]{P.E} [\mathit{allDiff}([\mathtt{I}_3,\mathtt{I}_4,\mathtt{I}_5,\mathtt{I}_6,\mathtt{I}_7,\mathtt{I}_8])]$$

*and then applies to remove three redundant bits in the underlying representation of each remaining integer (which is not equal to* $0, 6, 8$*):*

$$
\left[\begin{array}{l}
\mathit{allDiff}([ \\
\quad [1,\mathtt{i}_{3,2},\mathtt{i}_{3,3},\mathtt{i}_{3,4},\mathtt{i}_{3,5},\mathtt{i}_{3,7},\mathtt{i}_{3,7},0], \\
\quad [1,\mathtt{i}_{4,2},\mathtt{i}_{4,3},\mathtt{i}_{4,4},\mathtt{i}_{4,5},\mathtt{i}_{4,7},\mathtt{i}_{4,7},0], \\
\quad [1,\mathtt{i}_{5,2},\mathtt{i}_{5,3},\mathtt{i}_{5,4},\mathtt{i}_{5,5},\mathtt{i}_{5,7},\mathtt{i}_{5,7},0], \\
\quad [1,\mathtt{i}_{6,2},\mathtt{i}_{6,3},\mathtt{i}_{6,4},\mathtt{i}_{6,5},\mathtt{i}_{6,7},\mathtt{i}_{6,7},0], \\
\quad [1,\mathtt{i}_{7,2},\mathtt{i}_{7,3},\mathtt{i}_{7,4},\mathtt{i}_{7,5},\mathtt{i}_{7,7},\mathtt{i}_{7,7},0], \\
\quad [1,\mathtt{i}_{8,2},\mathtt{i}_{8,3},\mathtt{i}_{8,4},\mathtt{i}_{8,5},\mathtt{i}_{8,7},\mathtt{i}_{8,7},0]])
\end{array}\right]
\xrightarrow[\mathtt{allDifferent}]{P.E}
\left[\begin{array}{l}
\mathit{allDiff}([ \\
\quad [\mathtt{i}_{3,2},\mathtt{i}_{3,3},\mathtt{i}_{3,4},\mathtt{i}_{3,5},\mathtt{i}_{3,7}], \\
\quad [\mathtt{i}_{4,2},\mathtt{i}_{4,3},\mathtt{i}_{4,4},\mathtt{i}_{4,5},\mathtt{i}_{4,7}], \\
\quad [\mathtt{i}_{5,2},\mathtt{i}_{5,3},\mathtt{i}_{5,4},\mathtt{i}_{5,5},\mathtt{i}_{5,7}], \\
\quad [\mathtt{i}_{6,2},\mathtt{i}_{6,3},\mathtt{i}_{6,4},\mathtt{i}_{6,5},\mathtt{i}_{6,7}], \\
\quad [\mathtt{i}_{7,2},\mathtt{i}_{7,3},\mathtt{i}_{7,4},\mathtt{i}_{7,5},\mathtt{i}_{7,7}], \\
\quad [\mathtt{i}_{8,2},\mathtt{i}_{8,3},\mathtt{i}_{8,4},\mathtt{i}_{8,5},\mathtt{i}_{8,7}]])
\end{array}\right]
$$

**Example 15.** *Consider equi-propagation of constraint (c) from Figure 13 given $E_3$ from Example 14. The rules that apply derive from the decomposition of the* `int_array_plus` *constraint to it* `int_plus` *parts. These dictate that* $\mathtt{I}_3, \mathtt{I}_4, \mathtt{I}_5 \leq 5$*:*

$$
\boxed{\begin{array}{l}
\mathtt{E}_3 \\ \hline
\mathtt{I}_2 = [1,1,1,1,1,1,\neg\mathtt{i}_{1,7},\neg\mathtt{i}_{1,7}], \\
\mathtt{I}_3 = [1,\mathtt{i}_{3,2},\mathtt{i}_{3,3},\mathtt{i}_{3,4},\mathtt{i}_{3,5},\mathtt{i}_{3,7},\mathtt{i}_{3,7},0], \\
\mathtt{I}_4 = [1,\mathtt{i}_{4,2},\mathtt{i}_{4,3},\mathtt{i}_{4,4},\mathtt{i}_{4,5},\mathtt{i}_{4,7},\mathtt{i}_{4,7},0], \\
\mathtt{I}_5 = [1,\mathtt{i}_{5,2},\mathtt{i}_{5,3},\mathtt{i}_{5,4},\mathtt{i}_{5,5},\mathtt{i}_{5,7},\mathtt{i}_{5,7},0]
\end{array}}
\xrightarrow[\mathtt{int_array\ _plus}]{}
\boxed{\begin{array}{l}
\mathtt{E}_4 = \mathtt{E}_3 \cup \{\mathtt{i}_{3,7}{=}0, \mathtt{i}_{4,7}{=}0, \mathtt{i}_{5,7}{=}0\} \\ \hline
\mathtt{I}_2 = [1,1,1,1,1,1,\neg\mathtt{i}_{1,7},\neg\mathtt{i}_{1,7}], \\
\mathtt{I}_3 = [1,\mathtt{i}_{3,2},\mathtt{i}_{3,3},\mathtt{i}_{3,4},\mathtt{i}_{3,5},0,0,0], \\
\mathtt{I}_4 = [1,\mathtt{i}_{4,2},\mathtt{i}_{4,3},\mathtt{i}_{4,4},\mathtt{i}_{4,5},0,0,0], \\
\mathtt{I}_5 = [1,\mathtt{i}_{5,2},\mathtt{i}_{5,3},\mathtt{i}_{5,4},\mathtt{i}_{5,5},0,0,0]
\end{array}}
$$

*Applying partial evaluation simplifies the constraint as follows:*

$$
\left[\begin{array}{l}
\mathtt{int_array_plus}([ \\
\quad [1,1,1,1,1,1,\neg\mathtt{i}_{1,7},\neg\mathtt{i}_{1,7}], \\
\quad [1,\mathtt{i}_{3,2},\mathtt{i}_{3,3},\mathtt{i}_{3,4},\mathtt{i}_{3,5},0,0,0], \\
\quad [1,\mathtt{i}_{4,2},\mathtt{i}_{4,3},\mathtt{i}_{4,4},\mathtt{i}_{4,5},0,0,0], \\
\quad [1,\mathtt{i}_{5,2},\mathtt{i}_{5,3},\mathtt{i}_{5,4},\mathtt{i}_{5,5},0,0,0]],\ 14\ )
\end{array}\right]
\xrightarrow[\mathtt{int_array_plus}]{P.E}
\left[\begin{array}{l}
\mathtt{int_array_plus}([ \\
\quad [\neg\mathtt{i}_{1,7},\neg\mathtt{i}_{1,7}], \\
\quad [\mathtt{i}_{3,2},\mathtt{i}_{3,3},\mathtt{i}_{3,4},\mathtt{i}_{3,5}], \\
\quad [\mathtt{i}_{4,2},\mathtt{i}_{4,3},\mathtt{i}_{4,4},\mathtt{i}_{4,5}], \\
\quad [\mathtt{i}_{5,2},\mathtt{i}_{5,3},\mathtt{i}_{5,4},\mathtt{i}_{5,5}]],\ 5\ )
\end{array}\right]
$$

To summarize Examples 13–15 observe that in the initial constraint model 3 constraints about 8 integers are represented in 56 bits. After constraint simplification 2 constraints remain and the 8 integers are represented using 28 bits:

$$
\begin{array}{ll}
\mathtt{I}_1 = [1,1,1,1,1,1,\mathtt{i}_{1,7},\mathtt{i}_{1,7}] & \mathtt{I}_5 = [1,\mathtt{i}_{5,2},\mathtt{i}_{5,3},\mathtt{i}_{5,4},\mathtt{i}_{5,5},0,0,0] \\
\mathtt{I}_2 = [1,1,1,1,1,1,\neg\mathtt{i}_{1,7},\neg\mathtt{i}_{1,7}] & \mathtt{I}_6 = [1,\mathtt{i}_{6,2},\mathtt{i}_{6,3},\mathtt{i}_{6,4},\mathtt{i}_{6,5},\mathtt{i}_{6,7},\mathtt{i}_{6,7},0] \\
\mathtt{I}_3 = [1,\mathtt{i}_{3,2},\mathtt{i}_{3,3},\mathtt{i}_{3,4},\mathtt{i}_{3,5},0,0,0] & \mathtt{I}_7 = [1,\mathtt{i}_{7,2},\mathtt{i}_{7,3},\mathtt{i}_{7,4},\mathtt{i}_{7,5},\mathtt{i}_{7,7},\mathtt{i}_{7,7},0] \\
\mathtt{I}_4 = [1,\mathtt{i}_{4,2},\mathtt{i}_{4,3},\mathtt{i}_{4,4},\mathtt{i}_{4,5},0,0,0] & \mathtt{I}_8 = [1,\mathtt{i}_{8,2},\mathtt{i}_{8,3},\mathtt{i}_{8,4},\mathtt{i}_{8,5},\mathtt{i}_{8,7},\mathtt{i}_{8,7},0]
\end{array}
$$

## 6. Compiling Constraints with BEE

BEE (Ben-Gurion Equi-propagation Encoder) is a tool which applies to encode finite domain constraint models to CNF. BEE was first introduced by Metodi and Codish (2012). During the encoding process, BEE performs optimizations based on equi-propagation and partial evaluation to improve the quality of the target CNF. BEE is implemented in (SWI) Prolog and can be applied in conjunction with the CryptoMiniSAT solver (Soos, 2010) through a Prolog interface (Codish, Lagoon, & Stuckey, 2008). CryptoMiniSAT offers direct support for `xor` clauses, and BEE takes advantage of this feature. BEE can be downloaded (Metodi, 2012) where one can also find the examples from this paper and others.

The source language for the BEE compiler is also called BEE. It is a constraint modeling language similar to FlatZinc (Nethercote et al., 2007), but with a focus on a subset of the language relevant for finite domain constraint problems. Five of the constraint constructs in the BEE language are those introduced in Section 2.1. The full language is presented in Table 2.

In BEE Boolean constants "*true*" and "*false*" are viewed as (integer) values "1" and "0". Constraints are represented as (a list of) Prolog terms. Boolean and integer variables are represented as Prolog variables, which may be instantiated when simplifying constraints. in Table 2, `X` and `Xs` (possibly with subscripts) denote a literal (a Boolean variable or its negation) and a vector of literals, `I` (possibly with subscript) denotes an integer variable, and `c` (possibly with subscript) denotes an integer constant. On the right column of the table are brief explanations regarding the constraints. The table introduces 26 constraint templates.

Constraints (1-2) are about variable declarations: Booleans and integers. Constraint (3) expresses a Boolean as an integer value. Constraints (4-8) are about Boolean (and reified Boolean) statements. The special cases of Constraint (5) for $\texttt{bool_array_or}([\texttt{X}_1, \ldots, \texttt{X}_\texttt{n}])$ and $\texttt{bool_array_xor}([\texttt{X}_1, \ldots, \texttt{X}_\texttt{n}])$ facilitate the specification of clauses and of `xor` clauses (supported directly in the CryptoMiniSAT solver by Soos, 2010). Constraint (8) specifies that sorting a bit pair $[\texttt{X}_1, \texttt{X}_2]$ (decreasing order) results in the pair $[\texttt{X}_3, \texttt{X}_4]$. This is a basic building block for the construction of sorting networks (Batcher, 1968) used to encode cardinality (linear Boolean) constraints during compilation as described by Asín, Nieuwenhuis, Oliveras, and Rodríguez-Carbonell (2011) and by Codish and Zazon-Ivry (2010). Constraints (9-14) are about integer relations and operations. Constraints (15-20) are about linear (Boolean, Pseudo Boolean, and integer) operations. Constraints (21-26) are about lexical orderings of Boolean and integer arrays.

A main design choice of BEE is that all integer variables are represented in the order-encoding. So, BEE is suitable for problems in which the integer variables take small or medium sized values. The compilation of a constraint model to a CNF using BEE goes through three phases.

1. Unary bit-blasting: integer variables (and constants) are represented as bit vectors in the order-encoding.

2. Constraint simplification: three types of actions are applied: equi-propagation, partial evaluation, and decomposition of constraints. Simplification is applied repeatedly until no rule is applicable.

| | | |
|---|---|---|
| **Declaring Variables** | | |
| (1) | `new_bool(X)` | declare Boolean X |
| (2) | $\texttt{new_int}(\texttt{I}, \texttt{c}_1, \texttt{c}_2)$ | declare integer I, $c_1 \leq I \leq c_2$ |
| (3) | `bool2int(X,I)` | $(X \Leftrightarrow I = 1) \wedge (\neg X \Leftrightarrow I = 0)$ |
| **Boolean (reified) Statements** | | op $\in$ {or, and, xor, iff} |
| (4) | $\texttt{bool_eq}(X_1, X_2)$ or $\texttt{bool_eq}(X_1, -X_2)$ | $X_1 = X_2$ or $X_1 = -X_2$ |
| (5) | $\texttt{bool_array_op}([X_1, \ldots, X_n])$ | $X_1$ op $X_2 \cdots$ op $X_n$ |
| (6) | $\texttt{bool_array_op_reif}([X_1, \ldots, X_n], X)$ | $X_1$ op $X_2 \cdots$ op $X_n \Leftrightarrow X$ |
| (7) | $\texttt{bool_op_reif}(X_1, X_2, X)$ | $X_1$ op $X_2 \Leftrightarrow X$ |
| (8) | $\texttt{comparator}(X_1, X_2, X_3, X_4)$ | $\texttt{sort}([X_1, X_2]) = [X_3, X_4]$ |
| **Integer relations (reified)** | | rel $\in$ {leq, geq, eq, lt, gt, neq} |
| **and arithmetic** | op $\in$ {plus, times, div, mod, max, min}, | op′ $\in$ {plus, times, max, min} |
| (9) | $\texttt{int_rel}(I_1, I_2)$ | $I_1$ rel $I_2$ |
| (10) | $\texttt{int_rel_reif}(I_1, I_2, X)$ | $I_1$ rel $I_2 \Leftrightarrow X$ |
| (11) | $\texttt{int_array_allDiff}([I_1, \ldots, I_n])$ | $\bigwedge_{i<j} I_i \neq I_j$ |
| (12) | $\texttt{int_abs}(I_1, I)$ | $\lvert I_1 \rvert = I$ |
| (13) | $\texttt{int_op}(I_1, I_2, I)$ | $I_1$ op $I_2 = I$ |
| (14) | $\texttt{int_array_op}'([I_1, \ldots, I_n], I)$ | $I_1$ op′ $\cdots$ op′ $I_n = I$ |
| **Linear Constraints** | | rel$\in${leq, geq, eq, lt, gt} |
| (15) | $\texttt{bool_array_sum_rel}([X_1, \ldots, X_n], I)$ | $(\Sigma X_i)$ rel I |
| (16) | $\texttt{bool_array_pb_rel}([c_1, \ldots, c_n], [X_1, \ldots, X_n], I)$ | $(\Sigma c_i * X_i)$ rel I |
| (17) | $\texttt{bool_array_sum_modK}([X_1, \ldots, X_n], c, I)$ | $((\Sigma X_i) \bmod c) = I$ |
| (18) | $\texttt{int_array_sum_rel}([I_1, \ldots, I_n], I)$ | $(\Sigma I_i)$ rel I |
| (19) | $\texttt{int_array_lin_rel}([c_1, \ldots, c_n], [I_1, \ldots, I_n], I)$ | $(\Sigma c_i * I_i)$ rel I |
| (20) | $\texttt{int_array_sum_modK}([I_1, \ldots, I_n], c, I)$ | $((\Sigma I_i) \bmod c) = I$ |
| **Lexical Order** | | |
| (21) | $\texttt{bool_arrays_lex}(Xs_1, Xs_2)$ | $Xs_1$ precedes (leq) $Xs_2$ in the lex order |
| (22) | $\texttt{bool_arrays_lexLt}(Xs_1, Xs_2)$ | $Xs_1$ precedes (lt) $Xs_2$ in the lex order |
| (23) | $\texttt{bool_arrays_lex_reif}(Xs_1, Xs_2, X)$ | $X \Leftrightarrow Xs_1$ precedes (leq) $Xs_2$ in the lex order |
| (24) | $\texttt{bool_arrays_lexLt_reif}(Xs_1, Xs_2, X)$ | $X \Leftrightarrow Xs_1$ precedes (lt) $Xs_2$ in the lex order |
| (25) | $\texttt{int_arrays_lex}(Is_1, Is_2)$ | $Is_1$ precedes (leq) $Is_2$ in the lex order |
| (26) | $\texttt{int_arrays_lexLt}(Is_1, Is_2)$ | $Is_1$ precedes (lt) $Is_2$ in the lex order |

Table 2: Syntax of BEE Constraints.

3. CNF encoding: the best suited encoding technique is applied to the simplified constraints.

Bit-blasting and equi-propagation in BEE follow the general descriptions from Sections 2.4 and 3.1. Bit-blasting is implemented through Prolog unification. Each declaration of the form $\texttt{new_int}(I, c_1, c_2)$ triggers a unification $I = [1, \ldots, 1, X_{c_1+1}, \ldots, X_{c_2}]$ (to ease presentation we assume that integer variables are represented in a positive interval starting from 0 but there is no such limitation in practice as BEE also supports negatives integers). BEE applies ad-hoc equi-propagators as described in Section 4. When an equality of the form $X = L$ (between a variable and a literal or a constant) is detected, then equi-propagation is implemented by unifying $X$ and $L$. This unification applies to all occurrences of $X$ and in this sense "propagates" to other constraints involving $X$.

**Decomposition** is about replacing complex constraints (for example about arrays) with simpler constraints (for example about array elements). Consider, for instance, the constraint `int_array_plus(As,Sum)`. It is decomposed to a list of `int_plus` constraints applying a straightforward divide and conquer recursive definition. At the base case, if `As=[A]` then the constraint is replaced by a constraint of the form `int_eq(A,Sum)` which equates the bits of `A` and `Sum`, or if $\texttt{As} = [\texttt{A}_1, \texttt{A}_2]$ then it is replaced by $\texttt{int_plus}(\texttt{A}_1, \texttt{A}_2, \texttt{Sum})$. In the general case `As` is split into two halves, then constraints are generated to sum these halves, and then an additional `int_plus` constraint is introduced to sum the two sums.

| $c = \texttt{allDiff}^*([Z_1, Z_2, Z_3, \ldots, Z_n])$ | |
|---|---|
| if in E | then add in $\mu_{\texttt{c}}(\texttt{E})$ |
| $i \in dom(Z_1)$<br>$i \notin dom(Z_k)$ $(k > 1)$ | $Z_1 = i$ the authors |
| $\{i, j\} \cap dom(Z_k) = \emptyset$<br>$(k > 2)$ | $dom(Z_1) \subseteq \{i, j\}$<br>$dom(Z_2) \subseteq \{i, j\}$<br>$Z_1 \neq Z_2$<br>$Z_k \neq i, Z_k \neq j$<br>$(k > 2)$ |

Figure 14: Simplification rules for $\texttt{allDiff}^*$.

As another example, consider the $\texttt{int_plus}(\texttt{A}_1, \texttt{A}_2, \texttt{A})$ constraint. One approach, supported by BEE, decomposes the constraint as an odd-even merger (from the context of odd-even sorting networks) (Batcher, 1968). Here, the sorted sequences of bits $\texttt{A}_1$ and $\texttt{A}_2$ are merged to obtain their sum `A`. This results in a model with $O(n \log n)$ `comparator` constraints (and later in an encoding with $O(n \log n)$ clauses). Another approach, also supported in BEE, does not decompose the constraint but encodes it directly to a CNF of size $O(n^2)$, as in the context of so-called totalizers (Bailleux & Boufkhad, 2003). A hybrid approach, leaves the choice to BEE, depending on the size of the domains of the variables involved. Finally, we note that the user can configure BEE to fix the way it compiles this constraint (and others).

**CNF encoding** is the last phase in the compilation of a constraint model. Each of the remaining simplified (bit-blasted) constraints is encoded directly to a CNF. These encodings are standard and similar to those applied in various tools. The BEE encodings are similar to those applied in Sugar (Tamura et al., 2009).

### 6.1 The All-Different Constraint in BEE

The all-different constraint specifies that a set of integer variables take all different values from their specified domains. This constraint has received much attention in the literature (see for example the survey in van Hoeve, 2001). BEE provides special treatment for this constraint.

In many applications, all-different constraints are applied to model the special case when the constraint is about "permutation". Namely, when $[\texttt{I}_1, \ldots, \texttt{I}_\texttt{n}]$ are all different but may take precisely $n$ different values. BEE identifies this special case and applies two additional ad-hoc equi-propagation rules for this case. The table of Figure 14 illustrates these rules. We annotate the constraint with a "*" to emphasize that it has been detected that it is about permutation. The first rule is about the case when only one integer (assume $\texttt{Z}_1$) can take the value $i$. The second rule is about the case where all variables except two, assume $\texttt{Z}_1$, $\texttt{Z}_2$, cannot take two values, assume `i, j`. Now, because the constraint is about permutation, we can determine that $\mathtt{Z_1}$ and $\mathtt{Z_2}$ must take the two values $\mathtt{i}$ and $\mathtt{j}$. To illustrate the second rule consider the following example.

**Example 16.** *Consider a constraint* $\mathtt{allDiff}(\mathtt{I_1},\dots,\mathtt{I_5})$ *on 5 integer variables taking values in the interval* $[0,4]$ *(exactly 5 values) where* $E_0$ *specifies that* $\mathtt{I_3}$*,* $\mathtt{I_4}$ *and* $\mathtt{I_5}$ *cannot take the values 0 and 1. Therefore we introduce equations which restrict* $\mathtt{I_1}$ *and* $\mathtt{I_2}$ *to take the values 0 and 1, and the corresponding ad-hoc rule for permutation applies:*

$$\begin{array}{|l|}\hline \mathtt{E_0} = \left\{\begin{array}{c} x_{3,1}{=}1, x_{4,1}{=}1, \\ x_{5,1}{=}1, x_{3,2}{=}1, \\ x_{4,2}{=}1, x_{5,2}{=}1 \end{array}\right\} \\ \hline \mathtt{I_1} = [\mathtt{x_{1,1}},\dots,\mathtt{x_{1,4}}], \\ \mathtt{I_2} = [\mathtt{x_{2,1}},\dots,\mathtt{x_{2,4}}], \\ \mathtt{I_3} = [1,1,\mathtt{x_{3,3}},\mathtt{x_{3,4}}], \\ \mathtt{I_4} = [1,1,\mathtt{x_{4,3}},\mathtt{x_{4,4}}], \\ \mathtt{I_5} = [1,1,\mathtt{x_{5,3}},\mathtt{x_{5,4}}] \\ \hline \end{array} \quad \xrightarrow[\mathtt{allDiff^*}]{\begin{array}{l}\mathtt{dom(I_k)} \cap \{0,1\}{=}\emptyset \\ \mathtt{k} > 2\end{array}} \quad \begin{array}{|l|}\hline \mathtt{E_1} = \mathtt{E_0} \cup \left\{\begin{array}{c} x_{1,2}{=}0,\dots,x_{1,4}{=}0, \\ x_{2,2}{=}0,\dots,x_{2,4}{=}0 \\ x_{1,1}{=}\neg x_{2,1} \end{array}\right\} \\ \hline \mathtt{I_1} = [\mathtt{x_{1,1}},0,\dots,0], \\ \mathtt{I_2} = [\neg\mathtt{x_{1,1}},0,\dots,0], \\ \mathtt{I_3} = [1,1,\mathtt{x_{3,3}},\mathtt{x_{3,4}}], \\ \mathtt{I_4} = [1,1,\mathtt{x_{4,3}},\mathtt{x_{4,4}}], \\ \mathtt{I_5} = [1,1,\mathtt{x_{5,3}},\mathtt{x_{5,4}}] \\ \hline \end{array}$$

To facilitate the implementation of ad-hoc equi-propagation of all-different constraints, BEE adopts a dual representation for integer variables occurring in these constraints combining the order encoding and the, so-called, direct encoding. This is essentially the same as the encoding proposed by Gent and Nightingale (2004). When declaring an integer variable $\mathtt{I}$, the bit-blast in the order encoding applies the corresponding unification $\mathtt{I} = [\mathtt{x_1},\dots,\mathtt{x_n}]$. When encountering $\mathtt{I}$ in an $\mathtt{allDiff}$ constraint, an additional bit-blast introduces $\mathtt{I'} = [\mathtt{d_0},\dots,\mathtt{d_n}]$ in the direct encoding, and a channeling formula $\mathtt{channel}(\mathtt{I},\mathtt{I'})$ is introduced.

The direct encoding is a unary representation $\mathtt{I'} = [\mathtt{d_0},\dots,\mathtt{d_n}]$ where each bit $d_i$ is true if and only if $\mathtt{I'} = \mathtt{i}$. So, exactly one of the bits takes the value true. For example, the value 3 in the interval $[0,5]$ is represented in 6 bits as $[0,0,0,1,0,0]$. In the dual representation the following channeling formula captures the relation between the two representations of an integer variable $\mathtt{I} = [\mathtt{x_1},\dots,\mathtt{x_n}]$ and $\mathtt{I'} = [\mathtt{d_0},\dots,\mathtt{d_n}]$.

$$\mathtt{channel}([\mathtt{x_1},\dots,\mathtt{x_n}],[\mathtt{d_0},\dots,\mathtt{d_n}]) = \left(\begin{array}{c} d_0 = \neg x_1 \\ \wedge\ d_n = x_n \end{array}\right) \wedge \bigwedge_{\mathtt{i}=1}^{\mathtt{n}-1} (\mathtt{d_i} \leftrightarrow \mathtt{x_i} \wedge \neg\mathtt{x_{i+1}})$$

Consider an $\mathtt{allDiff}$ constraint about $m$ integer variables that can take different values between 0 and $n$. During constraint simplification, the $\mathtt{allDiff}([\mathtt{I_1},\dots,\mathtt{I_m}])$ constraint is viewed through its direct encoding as a bit matrix where each row consists of the bits $[\mathtt{d_{i0}},\dots,\mathtt{d_{in}}]$ for $\mathtt{I_i}$ in the direct encoding. The element $d_{ij}$ is true iff $I_i$ takes the value $j$. The $j^{th}$ column specifies which of the $I_i$ take the value $j$ and hence, at most one variable in a column may take the value true. This representation has one main advantage: in the direct encoding we can decompose $\mathtt{allDiff}([\mathtt{I_1},\dots,\mathtt{I_m}])$, to a conjunction of $n+1$ constraints, one for each column $0 \le j \le n$, of the form $\mathtt{bool_array_sum_leq}([\mathtt{d_{1j}},\dots,\mathtt{d_{mj}}],1)$, which is arc-consistent. As soon as $d_{i,j} = 1$ $(I_i = j)$ we have $d_{i,j'} = 0$ $(I_i \neq j')$ for all $j' \neq j$. In contrast in the order encoding alone the decomposition to $O(m^2)$ constraints $\left\{\ \mathtt{int_neq}(\mathtt{I_i},\mathtt{I_j}) \mid i < j\ \right\}$ is not arc-consistent. We illustrate the advantage of the dual encoding for the $\mathtt{allDiff}$ constraint in Section 8.1.

```
:- use_module(bee_compiler, [bCompile/2]).
:- use_module(sat_solver, [sat/1]).

solve(Instance, Solution) :-
    encode(Instance, Map, Constraints),
    bCompile(Constraints, CNF),
    sat(CNF),
    decode(Map, Solution).
```

Figure 15: A generic application of BEE.

# 7. Using BEE

A typical BEE application has the form depicted as Figure 15 where the predicate `solve/2` takes a problem `Instance` and provides a `Solution`. The specifics of the application are in the call to `encode/3` which given the `Instance` generates the `Constraints` that solve it together with a `Map` relating instance variables with constraint variables. The calls to `bCompile/2` and `sat/1` compile the constraints to a `CNF` and solve it applying a SAT solver. If the instance has a solution, the SAT solver binds the constraint variables accordingly. Then, the call to `decode/2`, using the `Map`, provides a `Solution` in terms of the `Instance` variables. The definitions of `encode/3` and `decode/2` are application dependent and provided by the user. The predicates `bCompile/2` and `sat/1` are part of the tool and provide the interface to BEE and the underlying SAT solver.

## 7.1 Example BEE Application: Magic Graph Labeling

We illustrate the application of BEE using Prolog as a modeling language to solve a graph labeling problem. Graph labeling is about finding an assignment of integers to the vertices and edges of a graph subject to certain conditions. Graph labellings were introduced in the 60's and hundreds of papers on a wide variety of related problems have been published since then. See for example the survey by Gallian (2011) with more than 1200 references. Graph labellings have many applications. For instance in radars, X-ray crystallography, coding theory, etc.

We focus here on the vertex-magic total labeling (VMTL) problem where one should find for the graph $G = (V, E)$ a labeling that is a one-to-one map $V \cup E \to \{1, 2, \ldots, |V| + |E|\}$ with the property that the sum of the labels of a vertex and its incident edges is a constant $K$ independent of the choice of vertex. A problem instance takes the form $vmtl(G, K)$ specifying the graph $G$ and a constant $K$. In the context of Figure 15, the query `solve(vmtl(G,K),Solution)` poses the question: "Does there exist a vmtl labeling for $G$ with magic constant $K$?" It binds `Solution` to indicate such a labeling if one exists, or to "unsat" otherwise. Figure 16 illustrates an example problem instance together with its solution.

Figure 17 illustrates a Prolog program that implements the `encode/3` predicate for the VMTL problem. The call to predicate `declareInts/4` introduces the constraints which declare the integer variables for each vertex and edge in the graph, and generates the map. The call to predicate `sumToK/5` introduces the constraints that require the sum of the labels for each vertex with its incident edges to equals $K$. The auxiliary predicate

| An Instance | The Graph | A Solution |
|---|---|---|
| $\texttt{Instance} = \texttt{vmtl}(\texttt{G},\texttt{K}),$<br>$\texttt{G} = (\texttt{V},\texttt{E}),$<br>$\texttt{V} = [1,2,3,4],$<br>$\texttt{E} = [(1,2),(1,3),$<br>$(2,3),(3,4)],$<br>$\texttt{K} = 14$ | 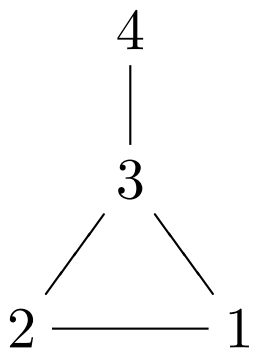 | $\begin{bmatrix} \texttt{V}_1 = 4, & \texttt{E}_{(1,2)} = 7, \\ \texttt{V}_2 = 5, & \texttt{E}_{(1,3)} = 3, \\ \texttt{V}_3 = 1, & \texttt{E}_{(2,3)} = 2, \\ \texttt{V}_4 = 6, & \texttt{E}_{(3,4)} = 8 \end{bmatrix}$ |

Figure 16: A VMTL instance with a solution.

```
encode(vmtl((Vs,Es),K),Map,Constraints):-
    append(Vs,Es,VEs), length(VEs,N),
    declareInts(VEs,N,Map,Constraints-Cs2),
    sumToK(Vs,Es,Map,K,Cs2-Cs3),
    getVars(VEs,Map,Vars),
    Cs3=[int_array_allDiff(Vars)].

declareInts([],_,_,Cs-Cs).
declareInts([ID|IDs],N,[(ID,X)|Map],[new_int(X,1,N)|CsH]-CsT):-
    declareInts(IDs,N,Map,CsH-CsT).

sumToK([],_,_,_,Cs-Cs).
sumToK([VID|Vs],Es,Map,K,[int_array_plus(Vars,K)|CsH]-CsT):-
    findall((X,Y),(member((X,Y),Es),(X=VID ; Y=VID)),EsIDs),
    getVars([VID|EsIDs],Map,Vars),
    sumToK(Vs,Es,Map,K,CsH-CsT).

getVars([],_,[]).
getVars([ID|IDs],Map,[Var|Vars]):-
    member((ID,Var),Map),
    getVars(IDs,Map,Vars).
```

Figure 17: encode/3 predicate for the VMTL application of BEE

| The Map | The Constraints |
|---|---|
| $((1,2),\texttt{E}_1),(1,\texttt{V}_1),$<br>$((1,3),\texttt{E}_2),(2,\texttt{V}_2),$<br>$((2,3),\texttt{E}_3),(3,\texttt{V}_3),$<br>$((3,4),\texttt{E}_4),(4,\texttt{V}_4)$ | new_int($\texttt{V}_1$, 1, 8), new_int($\texttt{E}_1$, 1, 8), int_array_plus([$\texttt{V}_1$, $\texttt{E}_1$, $\texttt{E}_2$], K),<br>new_int($\texttt{V}_2$, 1, 8), new_int($\texttt{E}_2$, 1, 8), int_array_plus([$\texttt{V}_2$, $\texttt{E}_1$, $\texttt{E}_3$], K),<br>new_int($\texttt{V}_3$, 1, 8), new_int($\texttt{E}_3$, 1, 8), int_array_plus([$\texttt{V}_3$, $\texttt{E}_2$, $\texttt{E}_3$, $\texttt{E}_4$], K),<br>new_int($\texttt{V}_4$, 1, 8), new_int($\texttt{E}_4$, 1, 8), int_array_plus([$\texttt{V}_4$, $\texttt{E}_4$], K),<br>new_int(K, 14, 14), allDiff([$\texttt{V}_1$, $\texttt{V}_2$, $\texttt{V}_3$, $\texttt{V}_4$, $\texttt{E}_1$, $\texttt{E}_2$, $\texttt{E}_3$, $\texttt{E}_4$]) |

Figure 18: A VMTL instance with the constraints and map generated by `encode/3`.

`getVars/3` receives a list of identifiers (vertices and edges) and extracts the corresponding list of integer variables from the map.

Given the VMTL instance from Figure 16, the call to predicate `encode/3` from Figure 17 generates the map and the constraints detailed in Figure 18.

Solving the constraints from Figure 18 binds the Map as follows, indicating a solution (in unary order encoding):

$$\mathtt{M} = \begin{bmatrix} (1,\ [1,1,1,1,0,0,0,0]), & ((1,2),\ [1,1,1,1,1,1,1,0]), \\ (2,\ [1,1,1,1,1,0,0,0]), & ((1,3),\ [1,1,1,0,0,0,0,0]), \\ (3,\ [1,0,0,0,0,0,0,0]), & ((2,3),\ [1,1,0,0,0,0,0,0]), \\ (4,\ [1,1,1,1,1,1,0,0]), & ((3,4),\ [1,1,1,1,1,1,1,1]) \end{bmatrix}$$

Using BEE to compile the constraints from Figure 18 generates a CNF which contains 301 clauses and 48 Boolean variables. Encoding the same set of constraints without applying simplification rules generates a larger CNF which contains 642 clauses and 97 Boolean variables.

In Section 8.3 we report that using BEE enables us to solve interesting instances of the VMTL problem not previously solvable by other techniques.

### 7.2 BumbleBEE

The BEE distribution includes also a command line solver, which we call BumbleBEE. BumbleBEE enables one to specify a BEE model in an input file where each line contains a single constraint from the model and the last line specifies the type of goal. BumbleBEE reads the input file, compiles the constraint model to CNF, solves the CNF using the embedded CryptoMiniSAT solver (Soos, 2010) and outputs a set of bindings to the declared variables in the model (or a message indicating that the constraints are not satisfiable). Figure 19 contains on the left the BumbleBEE input file for the VMTL instance from Figure 16 and on the right the BumbleBEE output, which is a solution for the constraint model. In the example, the last line of the input file specifies the goal to the solver. The options are:

1. `solve satisfy`: solve for a single satisfying assignment to the constraint model;

2. `solve satisfy(c)`: solve for (at most) $c$ satisfying assignments to the constraint model where $c$ is an integer value. When $c \leq 0$ this option will solve for all solutions.

3. `solve minimize(I)`: solve for a solution which minimizes the value of the integer variable $I$. The solver outputs the intermediate solutions (with decreasing values of $I$) encountered during the search for the minimum value of $I$.

4. `solve maximize(I)`: similar to minimize, but maximizes.

Further details and more examples can be found in the BEE distribution (Metodi & Codish, 2012).

## 8. Experiments

We report on our experience in applying BEE. To appreciate the ease in its use the reader is encouraged to view the example encodings available with the tool (Metodi & Codish, 2012). All experiments run on an Intel Core 2 Duo E8400 3.00GHz CPU with 4GB memory under Linux (Ubuntu lucid, kernel 2.6.32-24-generic). BEE is written in Prolog and run using SWI Prolog v6.0.2 64-bits. Comparisons with Sugar (v1.15.0) are based on the use of identical constraint models, apply the same SAT solver (CryptoMiniSAT v2.5.1), and run on the same machine. Times are reported in seconds.

| Content of BumbleBEE input file | BumbleBEE output |
|---|---|
| `new_int(V1, 1, 8)` | |
| `new_int(V2, 1, 8)` | |
| `new_int(V3, 1, 8)` | `V1 = 4` |
| `new_int(V4, 1, 8)` | `V2 = 5` |
| `new_int(E1, 1, 8)` | `V3 = 1` |
| `new_int(E2, 1, 8)` | `V4 = 6` |
| `new_int(E3, 1, 8)` | `E1 = 7` |
| `new_int(E4, 1, 8)` | `E2 = 3` |
| `int_array_plus([V1, E1, E2], 14)` | `E3 = 2` |
| `int_array_plus([V2, E1, E3], 14)` | `E4 = 8` |
| `int_array_plus([V3, E2, E3, E4], 14)` | `----------` |
| `int_array_plus([V4, E4], 14)` | `==========` |
| `int_array_allDiff([V1, V2, V3, V4, E1, E2, E3, E4])` | |
| `solve satisfy` | |

Figure 19: Solving VMTL instance using BumbleBEE.

### 8.1 Quasigroup Completion Problems

A Quasigroup Completion Problem (QCP) proposed by Gomes, Selman, and Crato (1997) as a constraint satisfaction benchmark, is given as an $n \times n$ board of integer variables (in the range $[1, n]$) in which some are assigned integer values. The task is to assign values to all variables, so that no column or row contains the same value twice. The constraint model is a conjunction of `allDiff` constraints. Ansótegui, del Val, Dotú, Fernández, and Manyà (2004) argue the advantage of the direct encoding for QCP.

We consider 15 instances from the 2008 CSP competition.[2] Table 3 considers three settings: BEE with its dual encoding for `allDiff` constraints, BEE using only the order encoding (equivalent to using `int_neq` constraints instead of `allDiff`), and Sugar. The table shows: the instance identifier ("sat" or "unsat"), compilation time (*comp*) in seconds, clauses in the encoding (*clauses*), variables in the encoding (*vars*), and SAT solving time (*SAT*) in seconds.

The results indicate that: (1) Application of BEE using the dual representation for `allDiff` is 38 times faster and produces 20 times fewer clauses (in average) than when using the order-encoding alone (despite the need to maintain two encodings); (2) Without the dual representation, solving encodings generated by BEE is only slightly faster than Sugar but BEE still generates CNF encodings 4 times smaller (on average) than those generated by Sugar. Observe that 3 instances are found unsatisfiable by BEE (indicated

2. `http://www.cril.univ-artois.fr/CPAI08/`. The competition instances are specified using binary disequalities, but here we use the model with `allDiff`.

| instance | | BEE (dual encoding) | | | | BEE (order encoding) | | | | Sugar | | |
|---|---|---|---|---|---|---|---|---|---|---|---|---|
| | | *comp* (sec) | *clauses* | *vars* | *SAT* (sec) | *comp* (sec) | *clauses* | *vars* | *SAT* (sec) | *clauses* | *vars* | *SAT* (sec) |
| 25-264-0 | sat | 0.23 | 6509 | 1317 | 0.33 | 0.36 | 33224 | 887 | 8.95 | 126733 | 10770 | 34.20 |
| 25-264-1 | sat | 0.20 | 7475 | 1508 | 3.29 | 0.30 | 34323 | 917 | 97.50 | 127222 | 10798 | 13.93 |
| 25-264-2 | sat | 0.21 | 6531 | 1329 | 0.07 | 0.30 | 35238 | 905 | 2.46 | 127062 | 10787 | 8.06 |
| 25-264-3 | sat | 0.21 | 6819 | 1374 | 0.83 | 0.29 | 32457 | 899 | 18.52 | 127757 | 10827 | 44.03 |
| 25-264-4 | sat | 0.21 | 7082 | 1431 | 0.34 | 0.29 | 32825 | 897 | 19.08 | 126777 | 10779 | 85.92 |
| 25-264-5 | sat | 0.21 | 7055 | 1431 | 3.12 | 0.30 | 33590 | 897 | 46.15 | 126973 | 10784 | 41.04 |
| 25-264-6 | sat | 0.21 | 7712 | 1551 | 0.34 | 0.33 | 39015 | 932 | 69.81 | 128354 | 10850 | 12.67 |
| 25-264-7 | sat | 0.21 | 7428 | 1496 | 0.13 | 0.30 | 36580 | 937 | 19.93 | 127106 | 10794 | 7.01 |
| 25-264-8 | sat | 0.21 | 6603 | 1335 | 0.18 | 0.27 | 31561 | 896 | 10.32 | 124153 | 10687 | 9.69 |
| 25-264-9 | sat | 0.21 | 6784 | 1350 | 0.19 | 0.27 | 35404 | 903 | 34.08 | 128423 | 10853 | 38.80 |
| 25-264-10 | unsat | 0.21 | 6491 | 1296 | 0.04 | 0.30 | 33321 | 930 | 10.92 | 126999 | 10785 | 57.75 |
| 25-264-11 | unsat | 0.12 | 1 | 0 | 0.00 | 0.28 | 37912 | 955 | 0.09 | 125373 | 10744 | 0.47 |
| 25-264-12 | unsat | 0.16 | 1 | 0 | 0.00 | 0.29 | 39135 | 984 | 0.08 | 127539 | 10815 | 0.57 |
| 25-264-13 | unsat | 0.12 | 1 | 0 | 0.00 | 0.29 | 35048 | 944 | 0.09 | 127026 | 10786 | 0.56 |
| 25-264-14 | unsat | 0.23 | 5984 | 1210 | 0.07 | 0.28 | 31093 | 885 | 11.60 | 126628 | 10771 | 15.93 |
| Total | | | | | 8.93 | | | | 349.58 | | | 370.63 |

Table 3: QCP results for $25 \times 25$ instances with 264 holes

by a CNF with a single clause and no variables). We comment that Sugar pre-processing times are higher than those of BEE and not indicated in the table.

## 8.2 Word Design for DNA

This is Problem 033 of CSPLib which seeks the largest parameter $n$, such that there exists a set $S$ of $n$ eight-letter words over the alphabet $\Sigma = \{A, C, G, T\}$ with the following properties: **(1)** Each word in $S$ has exactly 4 symbols from $\{C, G\}$; **(2)** Each pair of distinct words in $S$ differ in at least 4 positions; and **(3)** For every $x, y \in S$: $x^R$ (the reverse of $x$) and $y^C$ (the word obtained by replacing each $A$ by $T$, each $C$ by $G$, and vice versa) differ in at least 4 positions.

Mancini, Micaletto, Patrizi, and Cadoli (2008) provide a comparison of several state-of-the-art solvers applied to the DNA word problem with a variety of encoding techniques. Their best reported result is a solution with 87 DNA words, obtained in 554 seconds, using an OPL (van Hentenryck, 1999) model with lexicographic order to break symmetry. Frutos, Liu, Thiel, Sanner, Condon, Smith, and Corn (1997) present a strategy to solve this problem where the four letters are modeled by bit-pairs $[t, m]$. Each eight-letter word can then be viewed as the combination of a *"t-part"*, $[t_1, \ldots, t_8]$, which is a bit-vector, and a *"m-part"*, $[m_1, \ldots, m_8]$, also a bit-vector. The authors report a solution composed from two pairs of (t-part and m-part) sets[3] $[T_1, M_1]$ and $[T_2, M_2]$ where $|T_1| = 6$, $|M_1| = 16$, $|T_2| = 2$, $|M_2| = 6$. This forms a set $S$ with $(6 \times 16) + (2 \times 6) = 108$ DNA words. Marc van Dongen reports a larger solution with 112 words.[4]

Building on the approach described by Frutos et al. (1997), we pose conditions on sets of *"t-parts"* and *"m-parts"*, $T$ and $M$, so that their Cartesian product $S = T \times M$ will satisfy the requirements of the original problem. From the three conditions below, $T$ is required to satisfy (1′) and (2′), and $M$ is required to satisfy (2′) and (3′). For a set of

3. Their notions of t-part and m-part are slightly different than ours.
4. See `http://www.cs.st-andrews.ac.uk/~ianm/CSPLib/`.

bit-vectors $V$, the conditions are: **(1′)** Each bit-vector in $V$ sums to 4; **(2′)** Each pair of distinct bit-vectors in $V$ differ in at least 4 positions; and **(3′)** For each pair of bit-vectors (not necessarily distinct) $u, v \in V$, $u^R$ (the reverse of $u$) and $v^C$ (the complement of $v$) differ in at least 4 positions. This is equivalent to requiring that $(u^R)^C$ differs from $v$ in at least 4 positions.

It is this strategy that we model in our BEE encoding. An instance takes the form $\mathtt{dna}(\mathtt{n}_1, \mathtt{n}_2)$ signifying the numbers of bit-vectors, $n_1$ and $n_2$ in the sets $T$ and $M$. Without loss of generality, we impose, to remove symmetries, that $T$ and $M$ are lexicographically ordered. A solution is the Cartesian product $S = T \times M$.

Using BEE, we find, in a fraction of a second, sets of t-parts of size 14 and m-parts of size 8. This provides a solution of size $14 \times 8 = 112$ to the DNA word problem. Running Comet (v2.0.1) we find a 112 word solution in about 10 seconds using a model by Håkan Kjellerstrand.[5] Using BEE, we also prove that there does not exist a set of 15 t-parts (0.15 seconds), nor a set of 9 m-parts (4.47 seconds). These facts were unknown prior to BEE. Proving that there is no solution to the DNA word problem with more than 112 words, without the restriction to the two part t-m strategy, is still an open problem.

### 8.3 Vertex Magic Total Labeling

MacDougall, Miller, Slamin, and Wallis (2002) conjecture that the $n$ vertex complete graph, $K_n$, for $n \geq 5$ has a vertex magic total labeling with magic constants for a specific range of values of $k$, determined by $n$. This conjecture is proved correct for all odd $n$ and verified by brute force for $n = 6$. We address the cases for $n = 8$ and $n = 10$ which involve 15 instances (different values of $k$) for $n = 8$, and 23 (different values of $k$) for $n = 10$. Starting from the simple constraint model (illustrated by the example in Figure 16), we add additional constraints to exploit the fact that the graphs are symmetric: (1) We assume that the edge with the smallest label is $e_{1,2}$; (2) We assume that the labels of the edges incident to $v_1$ are ordered and hence introduce constraints $e_{1,2} < e_{1,3} < \cdots < e_{1,n}$; (3) We assume that the label of edge $e_{1,3}$ is smaller than the labels of the edges incident to $v_2$ (except $e_{1,2}$) and introduce constraints accordingly. In this setting BEE can solve all except 2 instances with a 4 hour timeout and Sugar can solve all except 4.

Table 4 gives results for the 10 hardest instances for $K_8$ the 20 hardest instances for $K_{10}$ with a 4 hour time-out. BEE compilation times are on the order of 0.5 sec/instance for $K_8$ and 2.5 sec/instance for $K_{10}$. Sugar encoding times are slightly larger. The instances are indicated by the magic constant, $k$; the columns for BEE and Sugar indicate SAT solving times (in seconds). The bottom two lines indicate average encoding sizes (numbers of clauses and variables).

The results indicate that the Sugar encodings are (in average) about 60% larger, while the average SAT solving time for the BEE encodings is about 2 times faster (average excluding instances where Sugar times-out).

To address the two VMTL instances not solvable using the BEE models described above ($K_{10}$ with magic labels 259 and 258), we partition the problem fixing the values of $e_{1,2}$ and $e_{1,3}$ and maintaining all of the other constraints. Analysis of the symmetry breaking constraints indicates that this results in 198 new instances for each of the two cases. The

5. See `http://www.hakank.org/comet/word_design_dna1.co`.

| instance | | BEE | Sugar |
|---|---|---|---|
| $K_8$ | $k$ | SAT (sec) | SAT (sec) |
| | 143 | 1.26 | 2.87 |
| | 142 | 10.14 | 1.62 |
| | 141 | 7.64 | 2.94 |
| | 140 | 14.68 | 6.46 |
| | 139 | 25.60 | 6.67 |
| | 138 | 12.99 | 2.80 |
| | 137 | 22.91 | 298.58 |
| | 136 | 14.46 | 251.82 |
| | 135 | 298.54 | 182.90 |
| | 134 | 331.80 | $\infty$ |
| Average CNF size: | | | |
| clauses | | 248000 | 402000 |
| vars | | 5688 | 9370 |

| instance | | BEE | Sugar |
|---|---|---|---|
| $K_{10}$ | $k$ | SAT (sec) | SAT (sec) |
| | 277 | 5.31 | 9.25 |
| | 276 | 7.11 | 9.91 |
| | 275 | 13.57 | 19.63 |
| | 274 | 4.93 | 9.24 |
| | 273 | 45.94 | 9.03 |
| | 272 | 22.74 | 86.45 |
| | 271 | 7.35 | 9.49 |
| | 270 | 6.03 | 55.94 |
| | 269 | 5.20 | 11.05 |
| | 268 | 94.44 | 424.89 |
| | 267 | 88.51 | 175.70 |
| | 266 | 229.80 | 247.56 |
| | 265 | 1335.31 | 259.45 |
| | 264 | 486.09 | 513.61 |
| | 263 | 236.68 | 648.43 |
| | 262 | 1843.70 | 6429.25 |
| | 261 | 2771.60 | 7872.76 |
| | 260 | 4873.99 | $\infty$ |
| | 259 | $\infty$ | $\infty$ |
| | 258 | $\infty$ | $\infty$ |
| Average CNF size: | | | |
| clauses | | 1229000 | 1966000 |
| vars | | 15529 | 25688 |

Table 4: VMTL results for $K_8$ and $K_{10}$ (4 hour timeout)

original VMTL instance is solved if any one of of these 198 instances is solved. So, we solve them in parallel. Fixing $e_{1,2}$ and $e_{1,3}$ "fuels" the compiler so the encodings are considerably smaller. The instance for $k = 259$ is solved in 1379.50 seconds where $e_{1,2} = 1$ and $e_{1,3} = 6$. The compilation time is 2.09 seconds and the encoding consists in just over 1 million clauses and 15 thousand variables.

To the best of our knowledge, the hard instances from this suite are beyond the reach of all previous approaches to program the search for magic labels. The SAT based approach presented by Jäger (2010) cannot handle these.[6] The comparison with Sugar indicates the impact of the compiler.

### 8.4 Balanced Incomplete Block Designs

This is Problem 028 of CSPlib (BIBD) where an instance is defined by a 5-tuple of positive integers $[v, b, r, k, \lambda]$ and requires to partition $v$ distinct objects into $b$ blocks such that each block contains $k$ different objects, exactly $r$ objects occur in each block, and every two distinct objects occur in exactly $\lambda$ blocks.

6. Personal communication (Gerold Jäger), March 2012.

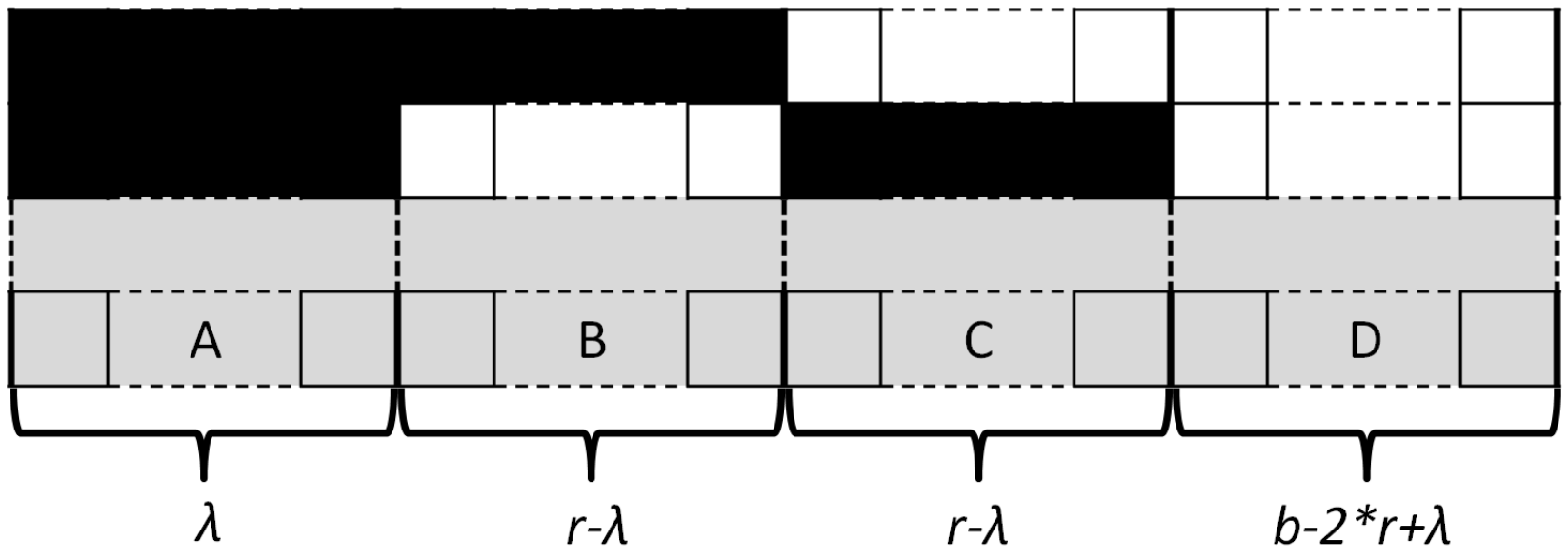

Figure 20: BIBD symmetry breaking.

The naive model for a BIBD instance $[v, b, r, k, \lambda]$ introduces the following constraints on a $v$ by $b$ Boolean incidence matrix: (1) exactly $r$ ones in each row, (2) exactly $k$ ones in each column, and (3) exactly $\lambda$ ones in each scalar product of two (different) rows.

This model does not contain a sufficient degree of information to trigger the equi-propagation process. In order to take advantage of the BEE simplifications we added symmetry breaking as described by Frisch, Jefferson, and Miguel (2004) and illustrated in Figure 20: Each row is viewed as sequence of four parts $A \ldots D$ with sizes $\lambda$, $(r - \lambda)$, $(r - \lambda)$, and $(b - 2r + \lambda)$. The first row is fixed by assigning parts $A$ and $B$ with ones (marked in black) and parts $C$ and $D$ with zeros (marked in white). The second row is fixed by assign parts $A$ and $C$ with ones (marked in black) and parts $B$ and $D$ with zeros (marked in white). For the third and all subsequent rows (marked in gray), the sum constraints are decomposed into summing each part $(A \ldots D)$ and then summing the results as follows: $A + B = \lambda$, $A + C = \lambda$, $C + D = r - \lambda$, and $B + D = r - \lambda$. This ensures that the row contains exactly $r$ ones and that the scalar product with the first (and second) row is $\lambda$. We denote this constraint model `SymB` (for symmetry breaking).

| instance | BEE (SymB) | | | Sugar (SymB) | | | SatELite (SymB) | | |
|---|---|---|---|---|---|---|---|---|---|
| $[v, b, r, k, \lambda]$ | *comp* (sec) | *clauses* | *SAT* (sec) | *comp* (sec) | *clauses* | *SAT* (sec) | *comp* (sec) | *clauses* | *SAT* (sec) |
| $[7, 420, 180, 3, 60]$ | 1.65 | 698579 | 1.73 | 12.01 | 2488136 | 13.24 | 1.67 | 802576 | 2.18 |
| $[7, 560, 240, 3, 80]$ | 3.73 | 1211941 | 13.60 | 11.74 | 2753113 | 36.43 | 2.73 | 1397188 | 5.18 |
| $[12, 132, 33, 3, 6]$ | 0.95 | 180238 | 0.73 | 83.37 | 1332241 | 7.09 | 1.18 | 184764 | 0.57 |
| $[15, 45, 24, 8, 12]$ | 0.51 | 116016 | 8.46 | 4.24 | 466086 | $\infty$ | 0.64 | 134146 | $\infty$ |
| $[15, 70, 14, 3, 2]$ | 0.56 | 81563 | 0.39 | 23.58 | 540089 | 1.87 | 1.02 | 79542 | 0.20 |
| $[16, 80, 15, 3, 2]$ | 0.81 | 109442 | 0.56 | 64.81 | 623773 | 2.26 | 1.14 | 105242 | 0.35 |
| $[19, 19, 9, 9, 4]$ | 0.23 | 39931 | 0.09 | 2.27 | 125976 | 0.49 | 0.4 | 44714 | 0.09 |
| $[19, 57, 9, 3, 1]$ | 0.34 | 113053 | 0.17 | $\infty$ | — | — | 10.45 | 111869 | 0.14 |
| $[21, 21, 5, 5, 1]$ | 0.02 | 0 | 0.00 | 31.91 | 3716 | 0.01 | 0.01 | 0 | 0.00 |
| $[25, 25, 9, 9, 3]$ | 0.64 | 92059 | 1.33 | 42.65 | 569007 | 8.52 | 1.01 | 97623 | 8.93 |
| $[25, 30, 6, 5, 1]$ | 0.10 | 24594 | 0.06 | 16.02 | 93388 | 0.42 | 1.2 | 23828 | 0.05 |
| Total (sec) | 36.66 | | | > 722.93 | | | > 219.14 | | |

Table 5: BIBD results (180 sec. timeout)

Table 5 shows results comparing BEE (compilation time, clauses in encoding, and SAT solving time) with Sugar using the `SymB` model. We also compare BEE with SatELite (Eén

& Biere, 2005), a CNF minimizer, where the input to SatELite is the CNF encoding for the SymB model generated by BEE without applying any simplifications. Here compilation time (*comp*) indicates the SatELite pre-processing time. The final row indicates the total of compilation and SAT solving time over the entire suite for each approach. In all cases time is measured in seconds.

This experiment indicates that BEE generates a significantly smaller CNF than Sugar which affects the SAT solving time. Moreover, the Sugar compilation time is extremely long. When comparing BEE with SatELite we can see that both output a CNF which is similar in size but as SatELite is applied on the entire CNF, for some instances its compilation time is significantly longer than its solving time.

| instance | BEE (SymB) | | Minion | | |
|---|---|---|---|---|---|
| $[v, b, r, k, \lambda]$ | *comp* | *SAT* | [M'06] | SymB | SymB$^+$ |
| $[7, 420, 180, 3, 60]$ | 1.65 | 1.73 | 0.54 | 1.36 | 0.42 |
| $[7, 560, 240, 3, 80]$ | 3.73 | 13.60 | 0.66 | 1.77 | 0.52 |
| $[12, 132, 33, 3, 6]$ | 0.95 | 0.73 | 5.51 | $\infty$ | 1.76 |
| $[15, 45, 24, 8, 12]$ | 0.51 | 8.46 | $\infty$ | $\infty$ | 75.87 |
| $[15, 70, 14, 3, 2]$ | 0.56 | 0.39 | 12.22 | 1.42 | 0.31 |
| $[16, 80, 15, 3, 2]$ | 0.81 | 0.56 | 107.43 | 13.40 | 0.35 |
| $[19, 19, 9, 9, 4]$ | 0.23 | 0.09 | 53.23 | 38.30 | 0.31 |
| $[19, 57, 9, 3, 1]$ | 0.34 | 0.17 | $\infty$ | 1.71 | 0.35 |
| $[21, 21, 5, 5, 1]$ | 0.02 | 0.00 | 1.26 | 0.67 | 0.15 |
| $[25, 25, 9, 9, 3]$ | 0.64 | 1.33 | $\infty$ | $\infty$ | 0.92 |
| $[25, 30, 6, 5, 1]$ | 0.10 | 0.06 | $\infty$ | 1.37 | 0.31 |
| Total | 36.66 | | >900.00 | >600.00 | 81.24 |

Table 6: BIBD results, comparison with Minion (times in seconds; 180 sec. timeout).

Table 6 shows results comparing BEE using the SymB model with the Minion constraint solver (Gent, Jefferson, & Miguel, 2006). We consider three different models for Minion: [M'06] indicates results using the BIBD model described by Gent et al. (2006), SymB uses the same model we use for the SAT approach, SymB$^+$, is an enhanced symmetry breaking model with all of the tricks applied also in the [M'06] model. For the columns with no timeouts we show total times (for BEE this includes compile time and SAT solving). Note that by using a clever modeling of the problem we have improved also the previous run-times for Minion.

This experiment indicates that BEE is significantly faster than Minion on its BIBD models ([M'06]). Only when tailoring our SymB model, does Minion becomes competitive with ours.

### 8.5 Combining BEE with SatELite

We now demonstrate the impact of combining BEE and SatELite. We describe experiments involving two of the benchmarks where SatELite is applied to simplify the output of BEE. The idea is to first apply the more powerful, but local, techniques, performed by BEE. This reduces the size of the CNF and is fast. Then we apply SatELite which takes global considerations on the CNF as a whole. We wish to determine if the smaller, simplified,

CNF is more amenable to further simplification using SatELite. The results indicate that although CNF size is slightly decreased, solving times are most often increased, sometimes drastically.

Tables 7 and 8 show our results. In both tables the four columns under the BEE heading indicate: BEE compilation time, size of the encoding (clauses and variables), and the subsequent SAT solving time. Similarly, the four columns under the Δ SatELite heading indicate the application of SatELite to the output of BEE: the SatELite processing time, the size of the resulting CNF (clauses and variables), and the subsequent SAT solving time. Table 7 illustrates the results for the BIBD benchmark of Section 8.4 and Table 8, the results for the 10 hardest VMTL instances for $K_8$ and for $K_{10}$ described in Section 8.3. Observe that applying SatELite to the output of BEE decreases the CNF size only slightly and does not improve the SAT solving time. In fact, to the contrary, in most cases it renders a CNF which takes more time to solve. In several cases, SAT solving time increases drastically to introduce a timeout.

| instance | BEE | | | | Δ SatELite | | | |
|---|---|---|---|---|---|---|---|---|
| $[v, b, r, k, \lambda]$ | *comp* (sec) | *clauses* | *vars* | *SAT* (sec) | *comp* (sec) | *clauses* | *vars* | *SAT* (sec) |
| $[7, 420, 180, 3, 60]$ | 1.65 | 698579 | 41399 | 1.73 | 1.88 | 696914 | 38749 | 3.41 |
| $[7, 560, 240, 3, 80]$ | 3.73 | 1211941 | 58445 | 13.60 | 3.14 | 1209788 | 54043 | 6.97 |
| $[12, 132, 33, 3, 6]$ | 0.95 | 180238 | 31947 | 0.73 | 1.20 | 179700 | 28351 | 0.91 |
| $[15, 45, 24, 8, 12]$ | 0.51 | 116016 | 19507 | 8.46 | 0.66 | 115938 | 17642 | $\infty$ |
| $[15, 70, 14, 3, 2]$ | 0.56 | 81563 | 19693 | 0.39 | 0.98 | 78630 | 15877 | 0.35 |
| $[16, 80, 15, 3, 2]$ | 0.81 | 109442 | 26223 | 0.56 | 1.13 | 104760 | 21116 | 0.50 |
| $[19, 19, 9, 9, 4]$ | 0.23 | 39931 | 9273 | 0.09 | 0.38 | 39805 | 7988 | 0.16 |
| $[19, 57, 9, 3, 1]$ | 0.34 | 113053 | 6576 | 0.17 | 12.49 | 112314 | 6230 | 0.37 |
| $[21, 21, 5, 5, 1]$ | 0.02 | 0 | 0 | 0.00 | 0.00 | 0 | 0 | 0.00 |
| $[25, 25, 9, 9, 3]$ | 0.64 | 92059 | 22098 | 1.33 | 0.97 | 91736 | 18540 | 10.34 |
| $[25, 30, 6, 5, 1]$ | 0.10 | 24594 | 2160 | 0.06 | 1.14 | 24028 | 1926 | 0.09 |

Table 7: BIBD results, BEE combined with SatELite (180 sec. timeout)

Our results demonstrate that the application of SatELite to remove redundancies from a CNF is often non-beneficial. Presumably the difference we see from our application of SatELite to other CNF benchmarks results from the fact that BEE produces highly optimized CNF output, while many CNF benchmarks have significant inefficiency in their original encoding. If BEE removes a variable from the CNF, then it also instantiates that variable, either to a constant or to an equivalent variable, and as such does not remove potential propagations from the encoding, as captured by Theorem 9.

## 9. Conclusion

There is a considerable body of work on CNF simplification techniques with a clear trade-off between amount of reduction achieved and invested time. Most of these approaches determine binary clauses implied by the CNF, which is certainly enough to determine Boolean equalities. The problem is that determining all binary clauses implied by the CNF is prohibitive when the SAT model may involve many (hundreds of) thousands of variables.

| instance | | BEE | | | | Δ SatELite | | | |
|---|---|---|---|---|---|---|---|---|---|
| | | *comp* (sec) | *clauses* | *vars* | *SAT* (sec) | *comp* (sec) | *clauses* | *vars* | *SAT* (sec) |
| $K_8$ | 143 | 0.51 | 248558 | 5724 | 1.26 | 2.60 | 248250 | 5452 | 0.98 |
| | 142 | 0.27 | 248414 | 5716 | 10.14 | 2.59 | 248107 | 5445 | 3.22 |
| | 141 | 0.20 | 248254 | 5708 | 7.64 | 2.59 | 247947 | 5437 | 32.81 |
| | 140 | 0.19 | 248078 | 5700 | 14.68 | 2.60 | 247771 | 5429 | 3.50 |
| | 139 | 0.18 | 247886 | 5692 | 25.6 | 2.59 | 247579 | 5421 | 6.18 |
| | 138 | 0.18 | 247678 | 5684 | 12.99 | 2.60 | 247371 | 5413 | 12.18 |
| | 137 | 0.18 | 247454 | 5676 | 22.91 | 2.59 | 247147 | 5405 | 77.16 |
| | 136 | 0.18 | 247214 | 5668 | 14.46 | 2.59 | 246907 | 5397 | 97.69 |
| | 135 | 0.18 | 246958 | 5660 | 298.54 | 2.58 | 246651 | 5389 | 705.48 |
| | 134 | 0.18 | 246686 | 5652 | 331.8 | 2.59 | 246379 | 5381 | $\infty$ |
| $K_{10}$ | 267 | 0.65 | 1228962 | 15529 | 88.51 | 3.02 | 1228368 | 14990 | 430.00 |
| | 266 | 0.65 | 1228660 | 15529 | 229.8 | 3.01 | 1228066 | 14990 | 259.55 |
| | 265 | 0.65 | 1228338 | 15529 | 1335.31 | 3.02 | 1227744 | 14990 | 540.48 |
| | 264 | 0.65 | 1227996 | 15529 | 486.09 | 3.02 | 1227402 | 14990 | 63.74 |
| | 263 | 0.65 | 1227634 | 15529 | 236.68 | 3.01 | 1227040 | 14990 | 1008.06 |
| | 262 | 0.65 | 1227252 | 15529 | 1843.7 | 3.02 | 1226658 | 14990 | 1916.73 |
| | 261 | 0.65 | 1226850 | 15529 | 2771.6 | 3.04 | 1226256 | 14990 | $\infty$ |
| | 260 | 0.65 | 1226428 | 15529 | 4873.99 | 3.02 | 1225834 | 14990 | $\infty$ |
| | 259 | 0.65 | 1225986 | 15529 | $\infty$ | 3.03 | 1225392 | 14990 | $\infty$ |
| | 258 | 0.65 | 1225524 | 15529 | $\infty$ | 3.01 | 1224930 | 14990 | $\infty$ |

Table 8: VTML results, BEE combined with SatELite (4 hour timeout)

Typically only some of the implied binary clauses are determined, such as those visible by unit propagation. The trade-off is regulated by the choice of the techniques applied to infer binary clauses, considering the power and cost. See for example the work of Eén and Biere (2005) and the references therein. There are also approaches (Li, 2003) that detect and use Boolean equalities during run-time, which are complementary to our approach.

In our approach, the beast is tamed by introducing a notion of locality. We do not consider the full CNF. Instead, by maintaining the original representation, a conjunction of constraints, each viewed as a Boolean formula, we can apply powerful reasoning techniques to separate parts of the model and maintain efficient pre-processing.

To this end, we introduce BEE, a compiler that follows this approach to encode finite domain constraints to CNF. Applying optimizations based on ad-hoc equi-propagation and partial evaluation rules on a high level view of the problem allows us to simplify the problem more aggressively than is possible with a CNF representation. The resulting CNF models can be significantly smaller than those resulting from straight translation.

It is well-understood that making a CNF smaller is not the ultimate goal: often smaller CNF's are harder to solve. Indeed, one often introduces redundancies to improve SAT encodings: so removing them is counterproductive. Our experience is that BEE reduces the size of an encoding in a way that is productive for the subsequent SAT solving. In particular, by removing variables that can be determined "at compile time" to be definitely equal (or definitely different) in any solution.

BEE uses ad-hoc equi-propagation and partial evaluation rules which keeps compilation times typically small (measured in seconds) even for instances which result in several millions of CNF clauses. And the reduction in SAT solving time can be larger by orders of magnitude. Hence, we believe that Boolean equi-propagation makes an important contribution to the encoding of CSPs to SAT.

BEE is currently tuned to represent integers in the order encoding. Ongoing work aims to extend BEE for binary and additional number representations such as mixed radix bases as considered by Eén and Sörensson (2006) and further by Codish, Fekete, Fuhs, and Schneider-Kamp (2011).

### Acknowledgments

We thank Vitaly Lagoon for the many insightful discussions concerning this research. NICTA is funded by the Australian Government as represented by the Department of Broadband, Communications and the Digital Economy and the Australian Research Council through the ICT Centre of Excellence Program.